\documentclass[journal]{IEEEtran}

\newcommand\ygk[1]{\textcolor{black}{#1}}

\ifCLASSINFOpdf

\else

\fi

\usepackage{url}
\usepackage{orcidlink}
\usepackage{multirow}
\usepackage[table,xcdraw]{xcolor}
\usepackage{booktabs}
\usepackage{amsmath,amssymb,amsfonts} 
\begin{document}

\title{Multi-view Structural Convolution Network for Domain-Invariant Point Cloud Recognition of Autonomous Vehicles}


\author{Younggun~Kim\orcidlink{0009-0005-3108-5304}, \ygk{Mohamed Abdel-Aty$^{\orcidlink{0000-0002-4838-1573}}$~\IEEEmembership{Senior Member,~IEEE}}, Beomsik~Cho\orcidlink{0009-0003-6883-7501}, Seonghoon~Ryoo\orcidlink{0009-0001-1943-7590},  and Soomok~Lee\IEEEauthorrefmark{2}\orcidlink{0000-0001-6633-997X}
        
\thanks{Younggun Kim and Mohamed Abdel-Aty are with Department of Civil Engineering, University of Central Florida, Florida 32816, USA (e-mail: younggun.kim@ucf.edu; m.aty@ucf.edu)}
\thanks{Beomsik Cho and Seonghoon Ryoo are with the Department of Data, Networks and AI, Ajou University, Suwon 16499, Korea. (e-mail: whqjatlr123@ajou.ac.kr; shryu0512@ajou.ac.kr)}

\thanks{\IEEEauthorrefmark{2}Soomok Lee is the corresponding author and is affiliated with the Department of Mobility Engineering and Department of Artificial Intelligence, Ajou University, Suwon 16499, Korea. (e-mail: soomoklee@ajou.ac.kr)}  
}

\markboth{Preprint}{Kim \MakeLowercase{\textit{et al.}}}

\maketitle

\begin{abstract}
Point cloud representation has recently become a research hotspot in the field of computer vision and has been utilized for autonomous vehicles. However, adapting deep learning networks for point cloud data recognition is challenging due to the variability in datasets and sensor technologies. This variability underscores the necessity for adaptive techniques to maintain accuracy under different conditions. In this paper, we present the Multi-View Structural Convolution Network (MSCN) designed for domain-invariant point cloud recognition. MSCN comprises Structural Convolution Layers (SCL) that extract local context geometric features from point clouds and Structural Aggregation Layers (SAL) that extract and aggregate both local and overall context features from point clouds. \ygk{Furthermore, MSCN enhances feature robustness by training with unseen domain point clouds generated from the source domain, enabling the model to acquire domain-invariant representations. Extensive cross-domain experiments demonstrate that MSCN achieves an average accuracy of \textbf{82.0\%}, surpassing the strong baseline PointTransformer by \textbf{15.8\%}, confirming its effectiveness under real-world domain shifts.} Our code is available at \url{https://github.com/MLMLab/MSCN}.
\end{abstract}


\maketitle

\section{Introduction}
\label{sec:introduction}
%
%
%
%

\IEEEPARstart{A}{utonomous} \ygk{vehicles (AVs) are envisioned as a transformative solution to major transportation challenges, including the mitigation of traffic accidents \cite{av_crash1, av_crash2, av_crash3}, congestion alleviation \cite{congestion1, congestion2}, and improved efficiency in transportation systems \cite{efficiency1}. Reliable perception is central to enabling these benefits, as AVs must accurately recognize and interpret their surroundings to ensure safety and efficiency. In this context, Light Detection and Ranging (LiDAR) has emerged as a particularly valuable sensor because it provides rich geometrical information, such as depth and structural details, that complements traditional camera-based perception. LiDAR-based point clouds provide accurate geometric representations of traffic environments, offering precise depth and structural information that is indispensable for robust AV perception and intelligent transportation systems (ITS).}

\ygk{With the advantages of LiDAR sensors, recent studies have increasingly leveraged LiDAR-centered perception within AVs for accurate 3D scene understanding with rich geometrical information and for robust perception under adverse weather conditions where camera-based systems often fail~\cite{fusion1, fusion2, fusion3, fusion4}. For example, recent work demonstrates that LiDAR maintains reliable detection and localization performance under adverse weather conditions such as rain, fog, and snow, highlighting its robustness where camera-based systems fail~\cite{fusion1}. Complementing this, Bhardwaj et al.~\cite{fusion2} propose a memristive associative learning circuit for fault-tolerant multi-sensor fusion, showing that incorporating LiDAR with other modalities ensures robust perception even when certain sensors degrade or fail in harsh environments.}

\begin{figure}[t]
\includegraphics[width=\columnwidth]{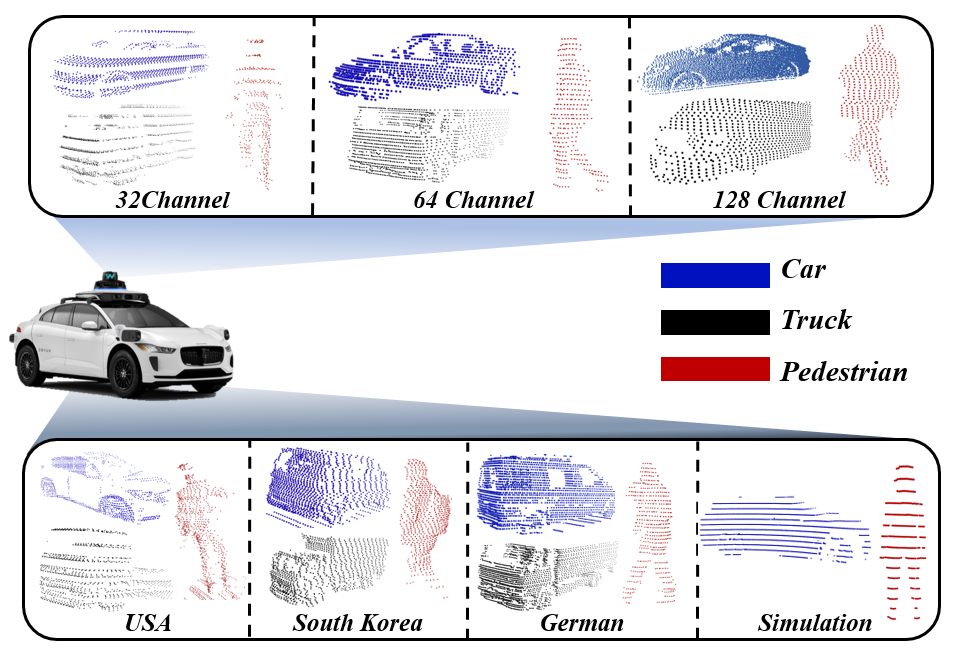}
\centering
\caption{\ygk{Illustration of LiDAR-based recognition's three key challenges, including different sensor configurations, geographic locations, and sim-to-real gap.}
}
\label{Fig1}
\end{figure}

\ygk{Despite the advances of LiDAR-based studies, prior research has largely overlooked the importance of achieving robust perception across heterogeneous LiDAR sensors. In practice, as illustrated in Figure~\ref{Fig1}, the properties of point clouds vary significantly depending on both the sensor configuration and the data acquisition domain. First, different sensor configurations, such as variations in the number of LiDAR channels, cause substantial discrepancies in point cloud density and resolution. For example, when models are trained on high-resolution LiDARs but deployed on cost-efficient low-resolution LiDARs, they often experience severe performance degradation. Since LiDAR configurations vary widely across manufacturers and even across research platforms, this mismatch presents a critical barrier to the scalable adoption of autonomous driving systems. Second, point cloud characteristics vary across geographic locations, as object appearances and structural patterns differ in distinct environments. Without accounting for such location-specific variations, autonomous vehicles risk misinterpreting unfamiliar object shapes, which can lead to failures in recognition and ultimately compromise driving safety. Moreover, the widely recognized sim-to-real gap poses another major challenge: while simulation environments enable the generation of abundant data and diverse accident scenarios, the point clouds they produce are typically uniform and noise-free. In contrast, real-world LiDAR data are often sparse and irregular, resulting in a pronounced distributional discrepancy that limits the generalization of models trained solely on simulation data.}

To address these challenges, we propose a novel deep learning model named Multi-view Structural Convolution Network (MSCN), which can extract features of 3D point clouds that are robust to domain changes in order to respond to arbitrary domain changes that frequently occur in point cloud recognition scenarios. The MSCN consists of a Structural Convolution Layer (SCL) and a Structural Aggregation Layer (SAL). In the SCL, structural information between points within the point cloud is captured in a local context, while in the SAL, structural information of the point cloud is captured in a global context and embedded into each point across multiple scales. By analyzing the structure of the point cloud from multiple perspectives, our MSCN can achieve highly sophisticated structural feature representation, ensuring stable performance despite domain changes.

In addition, point clouds are inherently irregular and exhibit significant variability depending on the LiDAR resolution and geographic location. To address these real-world irregularities and further improve the robustness of the model, we propose a novel approach for generating unseen point clouds based on source point clouds. Inspired by the Progressive Domain Expansion Network proposed for single-domain generalization in 2D images \cite{pden}, we extend this concept to 3D point cloud scenarios. By repeatedly generating virtual point clouds of new domains and training the model with these alongside the existing source point clouds, the model learns to generalize effectively to unseen point clouds encountered in real-world scenarios. This process enables the model to achieve greater invariance to domain changes and maintain consistent performance in various real-world conditions.

The main contributions of this study can be summarized in three folds as follows:

\begin{itemize}
\item We propose a Multi-view Structural Convolution Network (MSCN) that extracts structural features of point clouds from both specific local and overall contexts, strengthening feature robustness, especially for self-driving datasets including occluded point clouds and differing characteristics between different domains.

\item We propose a point cloud domain \ygk{expansion} framework by modifying the 2D Progressive Domain Expansion Network (PDEN)~\cite{pden} for effectively generating unseen domain point clouds, utilizing these generated virtual point clouds to train the model, thereby enhancing the MSCN’s robustness to domain changes.


\item We introduce a synthetic point cloud dataset to evaluate the consistent performance of our model in sim-to-real transitions. This dataset provides a controlled environment to validate the robustness and effectiveness of MSCN under simulated conditions. 

\item \ygk{We design progressively harder experimental settings to rigorously evaluate the domain-invariant recognition properties of MSCN:
(1) uniform synthetic point clouds with controlled geometric perturbations,
(2) indoor real-world point clouds with geometric variations,
(3) autonomous vehicle datasets with varying LiDAR channel configurations and geographic locations,
(4) sim-to-real domain shifts, and
(5) combined challenges of channel change and geometric perturbations. Through (1)–(2), we reveal that existing models remain highly sensitive to geometric transformations even under idealized synthetic and indoor environments.
Evaluations on autonomous driving datasets (3)–(5) further demonstrate MSCN’s robustness under both real-to-real and sim-to-real domain shifts, achieving an average cross-domain accuracy of 82.0\%, surpassing the strong baseline PointTransformer by 15.8\%.}

\end{itemize}

The subsequent sections of this paper are organized as follows. Section~\ref{sec:related} presents a comprehensive literature review of 3D point cloud encoding algorithms, domain adaptation methods, and domain generalization methods. Section~\ref{sec:MSCN} illuminates the proposed MSCN architecture. Section~\ref{sec:PDEN} shows a framework for progressively generating point clouds of unseen domains from a single domain point cloud and learning a feature extractor with the generated point clouds. In Section~\ref{sec:evaluation}, the paper presents an evaluation of the proposed network using various point cloud datasets acquired from synthetic datasets, real-world indoor dataset, and real-world self-driving vehicle datasets. Finally, Section~\ref{sec:conclusion} concludes the paper.

\section{Related Works}
\label{sec:related}

\subsection{3D point cloud deep learning networks: Encoding perspectives}

\subsubsection{Multi-view-based methods}
Considering that a collection of 2D views can offer significant information for 3D shape recognition compared to traditional manual feature extraction, several studies~\cite{MVCNN}, \cite{GVCNN}, \cite{RCPCNN}, \cite{MHBN} have explored recognizing 3D shapes from collections of rendered 2D images. Inspired by these works, Hamdi et al.~\cite{MVTN} proposed the Multi-View Transition Network (MVTN). This network includes a differentiable module that predicts the optimal viewpoint for a task-specific multi-view network, aiming to enhance 3D shape recognition by addressing limitations of previous multi-view methods. However, multi-view approaches still struggle to fully exploit the geometric relationships inherent in 3D data and extend their applicability to other 3D tasks, such as segmentation and reconstruction.

\subsubsection{Voxel-based methods}
To manage the unstructured nature of point clouds, some studies~\cite{VoxNet}, \cite{ShapeNet}, \cite{FPNN} have used voxel grids or voxelized 3D shapes as model inputs, enabling the application of standard 3D CNNs. Researchers like Riegler et al. \cite{OctNet} and Wang et al. \cite{O-CNN} have employed octree structures instead of voxels to improve computational efficiency and reduce memory consumption. Despite their focus on the relationships within 3D data and the grouping of point clouds, voxel-based methods often suffer from low classification efficiency due to the inherent sparsity and incomplete information in point clouds.

\subsubsection{Point cloud-based methods}
Recent research has increasingly focused on directly processing point clouds using deep learning techniques to fully utilize the information they contain. Aggregating local features is crucial in this process, as it captures rich information from point clouds. For example, PointNet, proposed by Qi et al.~\cite{pointnet}, is a pioneering study in this area. PointNet directly takes the point cloud as input, transforms it using the T-Net module, learns each point by sharing the full connection, and finally aggregates the features into global features through a max pooling function. However, PointNet only captures feature information from individual points and global points, without
considering the relational representation of adjacent points. This limitation prevents PointNet from effectively performing domain-invariant feature representation.

\subsection{3D point cloud deep learning networks: Local aggregation}

\subsubsection{Point-by-point methods}
Qi et al.~\cite{pointnet+} later proposed PointNet++ as an extension of PointNet. This method processes point clouds hierarchically, with each layer consisting of a sampling layer, a grouping layer, and a PointNet layer. The sampling layer identifies the centroid of the local neighborhood, the grouping layer constructs a subset of this neighborhood, and the PointNet layer captures the relationships between points within the local area. Despite its hierarchical processing, PointNet++ does not fully utilize the prior relationships between points. 
Ma et al.~\cite{pointMLP} observed that detailed local geometric information might not be essential for point cloud analysis, leading them to introduce PointMLP, a pure residual network without a complex local geometry extractor. Instead, it includes a lightweight geometric affine module, significantly improving inference speed. PointMLP achieved state-of-the-art results on both the ModelNet40 dataset~\cite{ModelNet40}, which is ideal for CAD-based analysis, and the ScanObjectNN dataset~\cite{ScanObjectNN}, which represents real indoor environments. However, this method may struggle with adaptability and consistent performance on real-world outdoor datasets, which are typically more irregular and sparse, due to its lack of precise local geometric information capture. 

\ygk{Beyond classification, several recent works have focused on robust feature representation for registration tasks, which are closely related to our goal of learning domain-invariant geometric features.
Ao et al.~\cite{Buffer} introduced BUFFER, a point-wise learner designed to predict keypoints and estimate point orientations, achieving a balance among accuracy, efficiency, and generalizability in registration. She et al.~\cite{Posdiffnet} proposed PosDiffNet, which leverages a graph-neural PDE based on Beltrami flow and integrates position embeddings into a Transformer-based ODE framework for multi-level correspondence learning, achieving state-of-the-art performance under large fields of view with perturbations.}

\subsubsection{Graph-based methods}
Wang et al.~\cite{dgcnn} proposed the Dynamic Graph CNN (DGCNN) for point cloud learning, introducing the edge convolutional (EdgeConv) network module. This module captures local geometric features of point clouds while maintaining arrangement invariance, underscoring the importance of local geometric features for 3D recognition tasks. However, the network is computationally intensive due to the complexity involved in handling dynamic graph structural modifications. The performance of DGCNN largely depends on the quality of the input graph. Suboptimal or perturbed graphs potentially compromise results, and perfecting graph construction methods is complex. 
To enhance the graphical relationships of point clouds in convolution, Lin et al.~\cite{gcn} proposed using 3D Graph Convolutional Networks to learn the geometric properties of point clouds. This approach considers geometric values and adjusts network parameters based on directional analysis, incorporating deformable graph kernels to handle translation, scale, and z-axis rotation. However, it does not sufficiently account for distances between point clouds, potentially missing detailed point-by-point interval nuances during parameter training. This limitation is particularly relevant for self-driving datasets, which feature small-scale, consistent patterns despite variations in LiDAR technology or underlying dataset changes. 
For better adaptation to self-driving datasets, Lee et al.~\cite{PanKyo} proposed 3D Structural Convolutional Networks, which include 3D convolution kernels to train 3D structural perspectives by combining cosine similarity and Euclidean distance terms. This approach enhances stability and consistent performance across varying point cloud datasets. However, it only captures the local geometric information of individual point clouds, potentially overlooking the overall structural properties of objects.

\subsection{Domain adaptation and domain generalization algorithms}

\subsubsection{Domain adaptation}
Recent studies~\cite{OOD1}, \cite{OOD2} have demonstrated that deep learning models are prone to significant performance degradation when evaluated on out-of-distribution (OOD) datasets, even with minor variations in the data generation process. These studies underscore the limited generalization capacity of deep learning models, which can severely impact their performance in real-world applications. Domain adaptation (DA)~\cite{DA1}, \cite{DA2}, \cite{DA3} has emerged as a strategy to tackle the OOD data challenge by utilizing a subset of data from the target domain. This approach enables a model trained on a source domain with different distributions to adapt to the target domain. The fundamental concept of DA is that, despite differences between source and target domains, common underlying structures or patterns can be leveraged to enhance model performance in the target domain. However, DA assumes the availability of either labeled or unlabeled data from the target domain for model adaptation, a condition not always met in real-world scenarios. Acquiring labeled data from the target domain can be prohibitively time-consuming, expensive, or impractical due to ethical or legal constraints. Additionally, the target domain's data distribution may remain unclear, complicating the application of conventional DA methods that rely on understanding this distribution.

\begin{figure*}[!t]
    \centerline{\includegraphics[width=\textwidth]{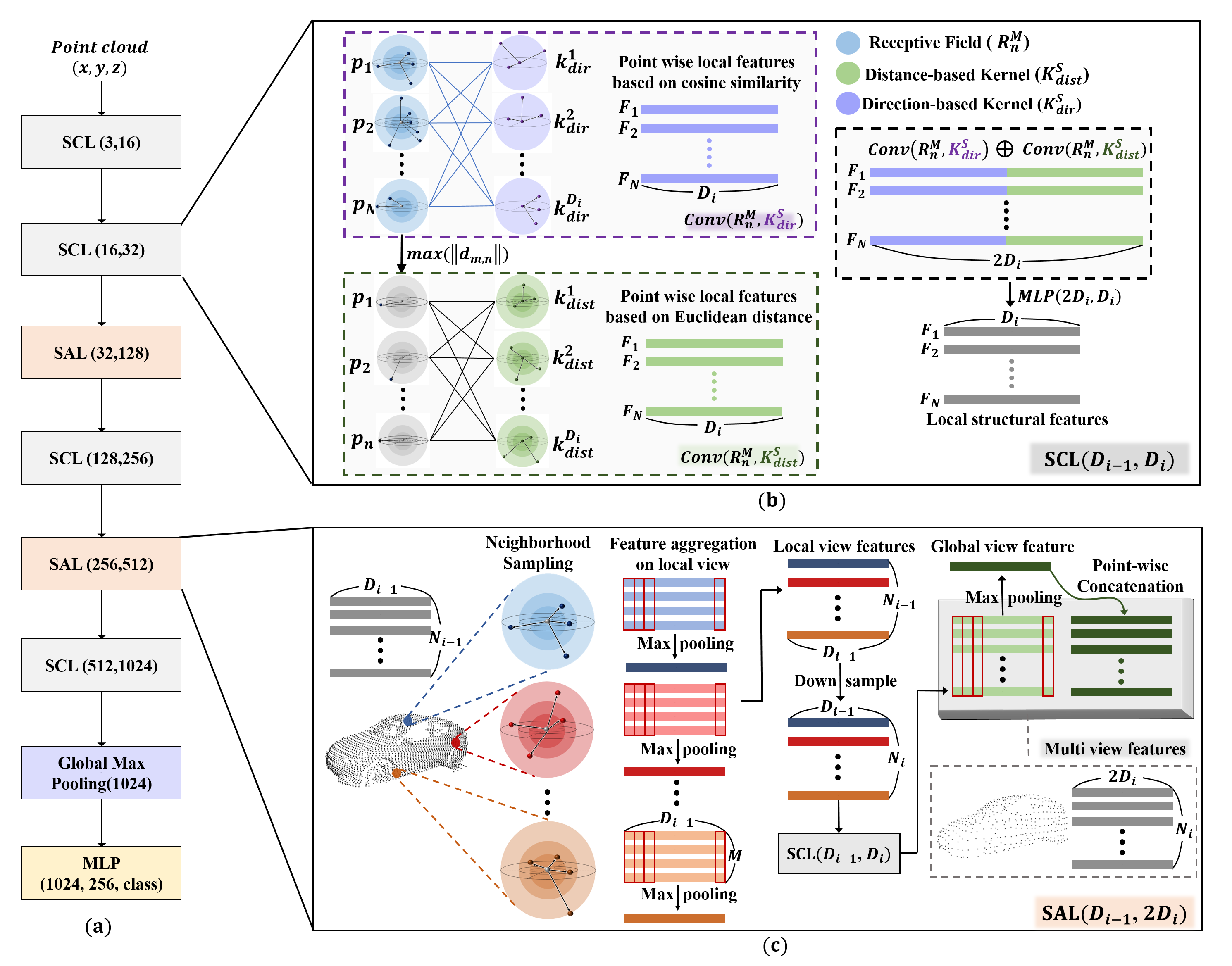}}
    \caption{The Multi-view Structural Convolution Network (MSCN) architecture for 3D point cloud classification comprises several components. (a) The MSCN includes feature extraction layers, specifically Structural Convolution Layers (SCL) and Structural Aggregation Layers (SAL), as well as Global Max Pooling and a multi-layer perceptron (MLP) for classification. (b) SCL is engineered to extract local features from each point. (c) The SAL is designed to combine local context features with overall context features.}
    \label{Fig2} 
\end{figure*}

\subsubsection{Single domain generalization}
To address the challenges associated with domain adaptation (DA), many domain generalization (DG) studies~\cite{DG1}, \cite{DG2}, \cite{DG3}, \cite{pden} particularly those focusing on single domain generalization, have been proposed. Single domain generalization is a particularly challenging form of model generalization where models are trained on a single domain and tested on unseen domains. A promising approach involves learning cross-domain invariant representations by expanding the coverage of the training domain. These methods typically synthesize additional samples in feature space to broaden the data distribution in the training set. 
Zhang et al.~\cite{DG1} observed that differences in medical images mainly arise in three aspects: image quality, image appearance, and spatial configuration. They augment these aspects through data augmentation. However, this method requires selecting the appropriate type and magnitude of augmentation based on the target domain, which can be challenging for other tasks. Volpi et al.~\cite{DG2} and Qiao et al.~\cite{DG3} synthesize additional data through adversarial learning to enhance model robustness. Nevertheless, this approach has limitations: the types of augmentation are relatively simple, and using too many adversarial examples during training can degrade classifier performance. To mitigate these issues, Li et al.~\cite{pden} proposed progressively generating multiple domains and learning a feature extractor using both source and generated data. The model employs contrastive learning to develop domain-invariant representations, effectively clustering each class to facilitate better decision boundaries and improve generalization. While this approach equips models to perform well in real-world scenarios, particularly on previously unseen domains, it is currently only applicable to domain generalization tasks involving well-structured 2D images.

\section{Multi-view Strucutral Convolution Network (MSCN)}
\label{sec:MSCN}

We focus on describing how to obtain domain-invariant features from the LiDAR point cloud and process it for classification. In the Structural Convolution Layer(SCL) section, we first explain how to extract the local context feature of the point cloud, which is achieved by the proposed structural convolution operation. Then, in the SAL section, we describe the feature extraction method from both local context and overall context of point cloud, making the feature representations more robust. Finally, we illustrate the architecture of point cloud classification.

\subsection{Structural Convolution Layer (SCL)}

A 3D point cloud object $P$ consists of $N$ points represented by $P = ({p_{n}|n = 1, 2, \ldots, N})$. Each $p_{n}$ represents a 3D coordinate within the three-dimensional space, thus belonging to the set $\mathbb{R}^{3}$. The structural convolution layer (SCL) captures the local structural and geometric characteristics associated with each point $p_{n}$. Figure~\ref{Fig2} (b) illustrates the i$^{th}$ structural convolution layer, named SCL($D_{i-1}, D_{i}$). First, SCL($D_{i-1}, D_{i}$) has receptive fields denoted ${R}^{M}_{n}$ to capture the relationships between the point $p_n$ and $M$ neighboring points. Specifically, the receptive field for the point $p_n$ is defined as:

\begin{equation}
R^{M}_ {n} = \{p_{n}, p_{m} | \forall p_{m} \in \mathcal{N}(p_{n},M)\},
\label{eq1}
\end{equation}
where $\mathcal{N}(p_{n},M)$ formally represents the $M$ nearest neighbor points of $ p_{n}$. Each $p_{n}$ has directional vectors within its receptive fields, defined as $d_{m, n} = p_{m} - p_{n}$, which will be used later for selecting the farthest neighbors among points $p_{m}$ in the receptive field and for performing the structural convolution operation. In $i^{th}$ SCL, all points in a point cloud have $D_{i-1}$ dimension feature vectors $F(p) \in \mathbb{R}^{D_{i-1}}$ and features contained within a receptive field can be expressed as $\{F (p_{n}), F (p_{m}) | \forall p_{m} \in \mathcal{N}(p_{n},M)\}$ where $F(p_{n}), F(p_{m}) \in \mathbb{R}^{D_{i-1}}$. To perform convolution directly on irregular points within the receptive field, SCL utilizes a Direction-based Kernel ($K^{S}_{dir}$) and Distance-based Kernel ($K^{S}_{dist}$) defined as follows:

\begin{equation}
K^{S}_{dir} = \{k^{1}_{dir}, k^{2}_{dir}, \ldots, k^{D_i}_{dir}\}
\label{eq2}
\end{equation}

\begin{equation}
K^{S}_{dist} = \{k^{1}_{dist}, k^{2}_{dist}, \ldots, k^{D_i}_{dist}\},
\label{eq3}
\end{equation}
where $S$ denotes the number of branches of the direction-based kernels and distance-based kernels. The branches of the directional kernel $k^{D_i}_{dir}$ are defined as $b^{1}_{dir}, b^{2}_{dir}, \ldots, b^{S}_{dir}$. Each branches consists of weight $w(b^{s}_{dir}) \in \mathbb{D}^{i},  s = 1,2, \ldots, S$ and a directional vector $b^{s}_{dir} \in \mathbb{R}^{3}$ indicating the spatial position of $w(b^{s}_{dir})$. To provide directionality to these branches, $k^{D_i}_{dir}$ needs to define the center position, which we set as $b^{c}_{dir}=(0, 0, 0)$. Since $b^{s}_{dir} = b^{s}_{dir} - b^{c}_{dir}$, learnable parameters in a direction-based kernel can be summarized as $ k^{D_i}_{dir} = \{w(b^{c}_{dir}), (w(b^{s}_{dir}), b^{s}_{dir}) | s= 1, 2, \ldots, S\}$. 
On the other hand, the distance-based kernel $k^{D_i}_{dist}$ contains branches defined as $b^{1}_{dist}, b^{2}_{dist}, \ldots, b^{S}_{dist}$ and these branches have weights defined as $w(b^{c}_{dist})$ and $w(b^{s}_{dist})$, indicating weight of the center point and branches respectively. Note that distance-based kernels are designed to analyze point relationships based on Euclidean distance, so the branches of $b_{dist}$ do not have directional vectors. Therefore, trainable parameters in a distance-based kernel can be summarized as $k^{D_i}_{dist} = \{w(b^{c}_{dist}), w(b^{s}_{dist}) | s= 1, 2, \ldots, S\}$.

Using the defined kernels, we perform convolution on irregular points within a receptive field of a point cloud. Specifically, we first assess the similarity between the features within the receptive field of $p_{n}$ (i.e. $F(p_{n})$, $F(p_{m})$ $\forall$ $p_{m} \in N(p_{n}, M)$, as specified in Equation~\ref{eq1} and the weight vectors of the directional kernel $K^{S}_{dir}$, centered around $b^{C}_{dir}$ with $S$ branches (namely, $w(b^{c}_{dir})$, $w(b^{s}_{dir})$ $\forall$ $s = 1, 2, \ldots, S$). We consider every pairing of $(p_{m}, b^{s}_{dir})$. Consequently, the direction-based convolution between a receptive field and a direction-based kernel can be described as:

\begin{equation}
\begin{split}
\mathrm{Conv}_{\mathrm{dir}}(R^{M}_{n}, K^{S}_{\mathrm{dir}})
&= \left\langle F(p_n),\, w\!\left(b^{c}_{\mathrm{dir}}\right) \right\rangle \\
&\quad + \sum_{s=1}^{S} \max_{m \in (1, M)}
   sim\!\left(p_{m},\, b^{s}_{dir}\right)
\end{split}
\label{eq4}
\end{equation}

where the symbol $\langle \cdot \rangle$ denotes the inner product operation, and the function $sim$ calculates the inner product between the features \(F(p_m)\) and the directional weights \(w(b^{s}_{dir})\), utilizing cosine similarity~\cite{gcn} to define this interaction:
\begin{equation}
sim(p_{m}, b^{s}_{dir}) = \langle F(p_{m}), w(b^{s}_{dir}) \rangle 
\frac{\langle d_{m,n}, b^{s}_{dir} \rangle}{||d_{m,n}||\,||b^{s}_{dir}||}
\label{eq5}
\end{equation}

The convolution operates between points of all receptive fields and the kernels since $k^{S}_{dir}$ has directional kernels $D_i$. Furthermore, to consider the impact of spatial relationships among neighboring points on structural characteristics, the Euclidean distances are calculated from the point $p_{n}$ to its most distant neighbor and then multiply these distances by the weights of the kernel $k^{D_i}_{dist}$ centered at $b^{c}_{dist}$ (specifically, $w(b^{c}_{dist}), w (b^{s}_{dist}), \forall s = 1, 2, \ldots, S$, $w(b^{c}_{dist}), w(b^{c}_{dist}) \in w_{dist}$). We consider all possible combinations of  $(p_{m}, b^{s}_{dist})$ to perform the distance-based convolution that merges the receptive fields with the kernels, expressed as:
\begin{equation}
\hspace{-2em}Conv_{dist}(R^{M}_{n}, K^{S}_{dist})= \sum_{s=1}^{S}w_{dist}\times\max\limits_{m \in (1,M)}(||d_{m, n}||) \hspace{-1em}
\label{eq6}
\end{equation}

Finally, by concatenating the outputs from the above convolution operations and passing them through an MLP, the structural convolution operation of the structural convolution layer $i^{th}$, denoted as $SCL(D_{i-1}, D_{i})$, is achieved as follows:

\begin{align}
SCL(D_{i-1}, D_{i}) & \nonumber \\
&\hspace{-6.8em} = \mathrm{MLP}\!\Big(
   \operatorname{Conv}_{\mathrm{dir}}(R^{M}_{n}, K^{S}_{\mathrm{dir}})
   \oplus
   \operatorname{Conv}_{\mathrm{dist}}(R^{M}_{n}, K^{S}_{\mathrm{dist}})
   \Big) \label{eq7}
\end{align}

\subsection{Structural Aggregation Layer (SAL)}

SAL $(D_{i-1}, D_i)$ is specially designed to capture the structural information of a point cloud from both local and global perspectives. As shown in Figure~\ref{Fig2} (c), SAL$(D_{i-1}, D_i)$ receives a point cloud from the previous layer, where all points have high-dimensional local features $D_{i-1}$ as input. This layer first performs neighborhood sampling to construct the receptive fields and then applies channel-wise max-pooling within these receptive fields to aggregate features $F(p)$, $\forall p \in R^{M}_{n}$, resulting in a coarse to fine feature representation at the local level. Next, a subset of $P$ is downsampled using a predetermined sampling rate $r$. These local view aggregation processes can be formulated as:

\begin{equation}
LocalViewAggregation(P_{in} , F_{in}) = (P_{out} , F_{out})
\label{eq8}
\end{equation}
where $P_{in} \in \mathbb{R}^{N_{i-1} \times 3}$, $P_{out} \in \mathbb{R}^{N_i \times 3}$, $F_{in} \in \mathbb{R}^{N_{i-1} \times D_{i-1}}$, $F_{out} \in \mathbb{R}^{N_i \times D_{i-1}}$, and $N_i = \frac{N_{i-1}}{r}$. This local feature aggregation operation allows us to learn multi-scale local context features, which are crucial factors for extracting local geometric information.

\begin{figure*}[!t]
    \centerline{\includegraphics[width=\textwidth]{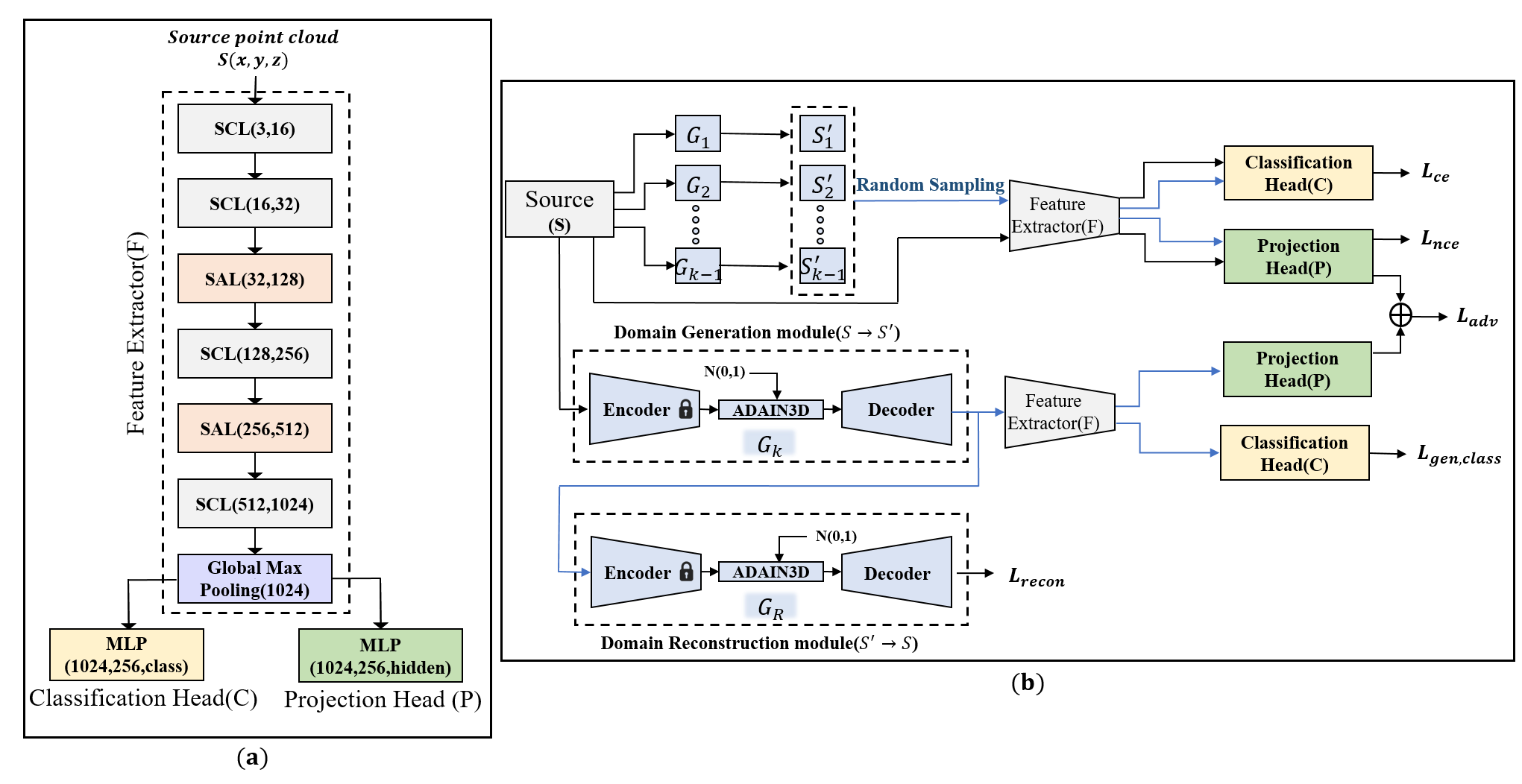}}
    \caption{The architecture for generating unseen domains in MSCN to learn more robust feature representations includes the following components. In (a), the modified MSCN is shown, consisting of a feature extractor (F), a classification head (C), and a projection head (P) with the projection head added for contrastive learning. In (b), the process of generating arbitrary unseen domains and training the MSCN with both source and generated data is illustrated. These generation and training processes are performed alternately.}
    \label{Fig3} 
\end{figure*}

The downsampled points $P_{out}$ and the fine features locally $F_{out}$ are used as input of SCL($D_{i-1}, D_{i}$) defined in Equation~\ref{eq7}, to extract geometric information at a lower scale. This results in effective geometric features across different scales, exhibiting both scale and shift invariance. The output features of SCL(${D_{i-1}, D_{i}}$) are summarized into a $D_i$ dimensional vector, called the global view feature, which represents the geometric information of the global context of a point cloud. This global view aggregation process can be formulated as:

\begin{equation}
GlobalViewAggregation(P_{in} , F_{in}) = F_{out},
\label{eq9}
\end{equation}
where $P_{in} \in \mathbb{R}^{N_{i} \times 3}$, $F_{in} \in \mathbb{R}^{N_{i} \times D_{i}}$, and $F_{out} \in \mathbb{R}^{D_{i}}$. By concatenating the global view feature with each row of output features from SCL$(D_{i-1}, D_{i})$, we can obtain multi-view features $F \in \mathbb{R}^{N_{i} \times 2D_{i}}$. This concatenation results in SAL($D_{i-1}, 2D_i$), the output of this layer, and serves as the input to the next structural convolution layer.

\subsection{Classification Process}

 For the purpose of point cloud classification, our framework consists of four structural convolution layers and two structural aggregation layers, as illustrated in Figure~\ref{Fig2} (a). As the point cloud data passes through these layers, it becomes more refined and is ultimately summarized into a 1024-dimensional vector using a max-pooling operation. Finally, this vector is fed into a multi-layer perceptron (MLP) for classification across the various classes.

\section{Progressive Domain Expansion Progress}
\label{sec:PDEN}
In section~\ref{sec:MSCN}, we explained how to obtain domain-invariant feature representations of point clouds by capturing geometric relationships between points. In this section, we focus on effectively generating unseen domains of 3d point clouds from specific domain point clouds, enabling MSCN to learn more robust feature representations.

\subsection{Generator composition}
To effectively generate unseen domains from a source domain, we adopt PDEN with several modifications. PDEN is originally designed for 2D image generation, which is structured into pixels, so we modify it to work with irregular point clouds. Especially, supplementing random sampling accommodates the sparse nature of point cloud patterns, making it easily adaptable to variations in LiDAR channels. 

Our goal is to generate $k$ arbitrary point cloud domains from a source point cloud domain. We define $k^{th}$ domain generator as $G_{k}$, which converts the original point cloud $x$ in a source domain $S$ into a synthetic point cloud $x^{'}_{k}$ in an unseen domain $S^{'}_{k}$. The generated point cloud and the unseen domain are formulated as:
\vspace{-0.1cm}
\begin{align}
\label{eq10}
 x^{'}_{k} = G_{k}(x, n), n \sim N (0, 1) \\
S^{'}_{k} = \{G_{k}(x_{i}, y_{i}) \mid (x_{i}, y_{i}) \in S_{k}\} \notag
\end{align}

where $x^{'} \in \mathbb{R}^{3}$ has the same semantic information as $x$, but $x^{'}$ and $x$ belong to different domains. We primarily use an autoencoder with AdaIN~\cite{adain} as a generator. However, since AdaIN was created for image transfer from a source domain to a specific target domain, we use two fully connected layers $L_{1}$ and $L_{2}$ along with Gaussian noise $N$ to learn appropriate parameters instead of a specific target domain. By modifying \ygk{AdaIN} to handle irregular point cloud features, our generator can be expressed as:

\begin{align}
\label{eq11}
G_{k}(x, n)= G_{D}(AdaIN_{3D}(G_{E}(x), n)), n \sim N(0, 1) \\
AdaIN_{3D}(z, n)= L_{1}\frac{z-\mu(z)}{\sigma(z)} + L_{2}  \notag
\end{align}

where $G_{E}$ and $G_{D}$ denote the encoder and decoder, respectively. Note that $\{G_{j}|, j=1, 2, \ldots, k-1\}$ represents an already trained generator set, which generates the arbitrary unseen domain set $\{S^{'}_{j}|, j=1, 2, \ldots, k-1\}$, with $G_{k}$ being the target generator to train. These generated domain sets will be used by MSCN to learn more domain-invariant feature representations.

\subsection{MSCN training with $\{S^{'}\}^{k-1}_{i=1}$}

As depicted in Figure~\ref{Fig3} (a), a projection head (P) is added to MSCN for contrastive learning, while the feature extractor (F) and classification head (C) remain the same as Figure~\ref{Fig2} (a). In the $k^{th}$ domain expansion environment, MSCN is trained with the updated dataset $S \cup \{S^{'}\}^{k-1}_{i=1}$, as shown in Figure~\ref{Fig3} (b). Specifically, given a batch $B= \{x_{i}, y_{i}\}^{2N}_{i=1}$ where $x_{i}$ is the source point cloud, $x^{'}_{i} \in \{S^{'}\}^{k-1}_{i=1}$ is the synthetic point cloud randomly sampled from the unseen domain set, and $y_{i}$ is the class label. MSCN learns domain-invariant feature representations by minimizing cross-entropy loss and InfoNCE loss~\cite{infoNCE} as follows:

\begin{align}
\label{eq12}
L_{ce}(x_{i}, y_{i}) &= \min\limits_{F,C} -\sum_{m=1}^{M} y^{m}_{i} \log(C(F(x^{m}_{i}))) \\
L_{nce}(z_i, z^{+}_{i}) &= \min\limits_{F,C} -\log\left(\frac{\exp(z_{i}\cdot z^{+}_{i})}{\sum^{2N}_{j=1, j \ne i} \exp(z_{i}\cdot z_{j})}\right) \notag
\end{align}

where $M$ is the number of label, $y^{m}_{i}$ is the $m^{th}$ label of $y_{i}$, $z_{i}= P(F(x_{i}))$, and $z^{+}_{i}= P(F(x^{'}_{i}))$. Total losses to train MSCN, named $L_{src}$, can be summarized as follow:

\begin{equation}
\label{eq13}
L_{src}= L_{ce}(x_{i}, y_{i}) +  L_{nce}(z_i, z^{+}_{i})
\end{equation}

\subsection{Unseen Domain $S^{'}_{k}$ Generation}

To effectively and safely generate domains, we compose the unseen domain generation process as shown in Figure~\ref{Fig3} (b). The source domain point cloud $x$ is converted to an unseen domain point cloud $x^{'}_{k}$ in the domain generation module $G_{k}$, which is then used as input for both the MSCN and the reconstruction module $G_{R}$. The primary goal of the reconstruction module is to reconstruct $x^{'}_{k}$ to $x_i$ to ensure the safety of the decoder $G_{D}$ and $AdaIN_{3D}$ in $G_{k}$. This is achieved by minimizing reconstruction loss defined as:

\begin{equation}
\label{eq14}
L_{recon}= \min\limits_{G_{k},G_{R}} ||x-G_{R}(G_{k}(x, n))||_{2}
\end{equation}

Secondly, MSCN helps $G_{k}$ generate unseen domains more reliably by ensuring that $x^{'}_{k}$ is of the same class as $x$. This is achieved by minimizing the loss of cross-entropy at the head of the class $C$, formulated as:

\begin{equation}
\label{eq15}
\hspace{-0.8em} 
L_{\text{gen, ce}} = \min\limits_{G_{k},F,C} L_{ce}\!\left(C(F(G_{k}(x,n))), y\right), 
 n \sim N(0,1)
\end{equation}

On the other hand, to effectively generate unseen domain, we employ adversarial learning. The generator $G_{k}$ is trained to maximize the InfoNCE loss while MSCN is trained to minimize the loss. Through adversarial training, $G_{k}$ will generate unseen domains from which MSCN cannot extract domain-shared representations, thereby enhancing MSCN's ability to extract cross-domain invariant representations. The adversarial loss is defined as:

\begin{align}
\label{eq16}
L_{adv} &= \min\limits_{G_{K}} -L^{*}_{nce}(P(F(x)), P(F(G_{k}(x,n)))) \nonumber \\
&\quad + \min\limits_{F,P} -L_{nce}(P(F(x)), P(F(G_{k}(x,n))))
\end{align}

where $L^{*}_{nce}$ indicates the modified InfoNCE loss for the adversarial loss to converge, and $L^{*}_{nce}$ is defined as:

\begin{equation}
\label{eq17}
L^{*}_{nce}(z_i, z^{+}_{i})= \sum_{i}^{2N}(1-log\frac{exp(z_{i}\cdot z^{+}_{i})}{\sum^{2N}_{j=1, j \ne i} exp(z_{i}\cdot z_{j})})
\end{equation}

Moreover, we use an additional loss function to encourage $G_{k}$ to generate more diverse samples.

\begin{equation}
\label{eq18}
L_{div}= \min\limits_{G_{K}} ||G_{k}(x, n_{1}), G_{k}(x, n_{2})||_{2}
\end{equation}

Finally, we can effectively and safely generate $k^{th}$ unseen domain $S^{'}_{k}$ by minimizing the total loss, named $L_{unseen}$, defined as:

\begin{equation}
\label{eq19}
L_{unseen}= L_{recon}+L_{gen,ce}+L_{adv}+L_{div}
\end{equation}

\section{Evaluation}
\label{sec:evaluation}

To evaluate the domain-invariant capabilities of the proposed MSCN, we conducted comprehensive experiments using both synthetic datasets and real-world datasets collected from self-driving vehicles. The synthetic dataset experiments aim to verify MSCN’s invariance to geometric transformations, including translation and scaling. These tests provide insights into the model's ability to maintain performance under controlled yet challenging conditions.

In the real-world dataset experiments, we evaluated MSCN’s robustness in scenarios where the LiDAR resolution differs between training and testing environments. This setup simulates practical conditions in autonomous driving, where sensor configurations may vary. Furthermore, we assessed MSCN’s performance under severe domain shifts, combining both dataset-specific variations and additional transformations such as rotation and translation. Even in these harsh conditions, our MSCN consistently demonstrates robust performance, highlighting its effectiveness in domain-adaptive tasks.

\subsection{Datasets \& Implementation Detail}

\subsubsection{Synthetic Datasets} We conducted experiments using ShapeNetPart~\cite{shapenet} and ModelNet40~\cite{ModelNet40} datasets. For the ShapeNetPart dataset, we evaluated the proposed MSCN on the part segmentation task, while for the ModelNet40 dataset, we performed object classification. Both experiments followed the standard training settings commonly used in previous work. In addition, specific experiments are conducted comparing MSCN with PointMLP~\cite{pointMLP} and PointTransformer~\cite{transformer} to demonstrate MSCN’s invariance to geometric transformations such as translation and scaling. For these experiments, we fixed the number of training epochs for all models to 30 and used a consistent point cloud size of 1024 points. These settings ensured a fair comparison across models and highlighted MSCN's robustness under varying geometric transformations.

\subsubsection{Real-world Indoor Dataset} ScanObjectNN~\cite{ScanObjectNN} is a challenging dataset of 3D real-world object point clouds, comprising approximately 15,000 instances categorized into 15 classes. It includes three commonly used variants—OBJ BG, OBJ ONLY, and PB T50 RS—that progressively increase in difficulty. The experimental setup adheres to the standard training configurations typically employed in previous studies.

\subsubsection{Real-World Self-Driving Vehicle Datasets} We conducted experiments using three real-world point cloud datasets, namely KITTI~\cite{KITTI}, PanKyo~\cite{PanKyo}, and nuScenes~\cite{nuScene}, as well as a synthetic dataset from the MORAI simulator~\cite{simulation}. These datasets, accessible through the following link \footnote{https://sites.google.com/site/cvsmlee/dataset}, originate from various countries and manufacturers, making them crucial for evaluating the inference capacity of the model and the robustness to domain shifts.

Each dataset has a different number of channels. Specifically, nuScenes and the synthetic dataset have 32 channels, KITTI has 64 channels, and PanKyo has 128 channels. These variations help to assess the robustness of the methods. We set the number of classes for the realistic datasets to three, which include car, truck, and pedestrian, while the synthetic datasets have two classes including car and pedestrian.

Comparisons between our methods are conducted, including MSCN and MSCN$^{\dag}$, and precious methods including PointNet~\cite{pointnet}, PointNeXt~\cite{pointNext}, PointMLP~\cite{pointMLP}, DGCNN~\cite{dgcnn}, GCN~\cite{gcn}, SCN~\cite{PanKyo}, PointTransformer~\cite{transformer}, and PointMamba~\cite{pointmamba}. PointMamba was first pre-trained on ShapeNet~\cite{shapenet} and then trained on KITTI, PanKyo, and nuScenes for the classification task. The \ygk{networks} were \ygk{evaluated} through the a standard accuracy metric.

To prevent overfitting, the number of training epochs for all methods, except MSCN$^{\dag}$, is limited to 10. MSCN$^{\dag}$ indicates that applying the unseen domain generation methods to MSCN. For MSCN$^{\dag}$, a new domain is generated every 15 training cycles and this process is repeated 20 times, resulting in a total of 300 epochs.

Furthermore, we use a pre-trained MSCN as the feature extractor in MSCN$^{\dag}$ and pre-trained SCL as the generator encoders in MSCN$^{\dag}$. This approach ensures the appropriate preservation of features from the source dataset during the generation of unseen domains. The generator encoders are not updated during training.

We set the batch size to 16 and the learning rate to 0.0004 for all methods. The number of branches ($S$) is set to 1 and the number of neighbors ($M$) is set to 3 for all graph-based methods within receptive fields, including GCN, SCN, MSCN, and MSCN$^{\dag}$. Both DGCNN and point-by-point methods such as PointNet, PointNeXt, and PointMLP require point cloud normalization, so we conducted point cloud normalization for these methods. All experiments were carried out on a workstation with a deca-core Intel Xeon Silver 4210R CPU clocked at 2.4 GHz and an NVIDIA RTX A5000 GPU with 24GB of memory.

\subsection{Evaluation on Synthetic Datasets: Robustness to geometric deformation}

\begin{table}[!t]
\caption{Object part segmentation results on ShapeNetPart~\cite{ShapeNet}, grouped by dataset. mIOU values are reported for each method and dataset.}
\centering
\begin{tabular}{c|c|c}
\hline
\textbf{Method} & \textbf{\#points} & \textbf{mIOU (\%)} \\ \hline
     PointCNN~\cite{PointCNN}       & 2048       & 86.1 \\ 
     PointNet++~\cite{pointnet+}      & 2048        & 85.1 \\ 
     SO-Net~\cite{SO-Net}           & \ygk{2048}       & \ygk{84.9} \\ 
     SpiderCNN~\cite{SpiderCNN}       & 2048       & 85.3 \\ 
     PointConv~\cite{PointConv}      & 2048        & 85.7 \\ 
     KPConv~\cite{KPConv}         & 2048        & 86.4 \\ 
     CRFConv~\cite{CRFConv}      & 2048        & 85.5 \\ 
     DensePoint~\cite{DensePoint}        & 2048      & 86.4 \\ 
     RS-CNN  w/o vot.~\cite{RS-CNN}  & 2048       & 85.8 \\ 
     RS-CNN  w/ vot.~\cite{RS-CNN}    & 2048      & 86.2 \\ 
     PointASNL~\cite{PointASNL}        & 2048       & 86.1 \\ 
     PointNet~\cite{pointnet}      & 2048       & 83.7 \\ 
     DGCNN~\cite{dgcnn}       & 2048       & 85.1 \\ 
     PointMamba~\cite{pointmamba}       &   2048    & 86.2 \\
     PointMLP~\cite{pointMLP}        & 2048       & 86.1 \\
     PointTransformer~\cite{transformer}       & 2048      & \textbf{86.6} \\ 
      \hline
     MSCN(Ours)    & \ygk{2048}   & \ygk{85.0} \\ \hline
\end{tabular}
\label{table:ShapeNet}
\end{table}

\begin{figure*}[!t]
    \centerline{\includegraphics[width=\textwidth]{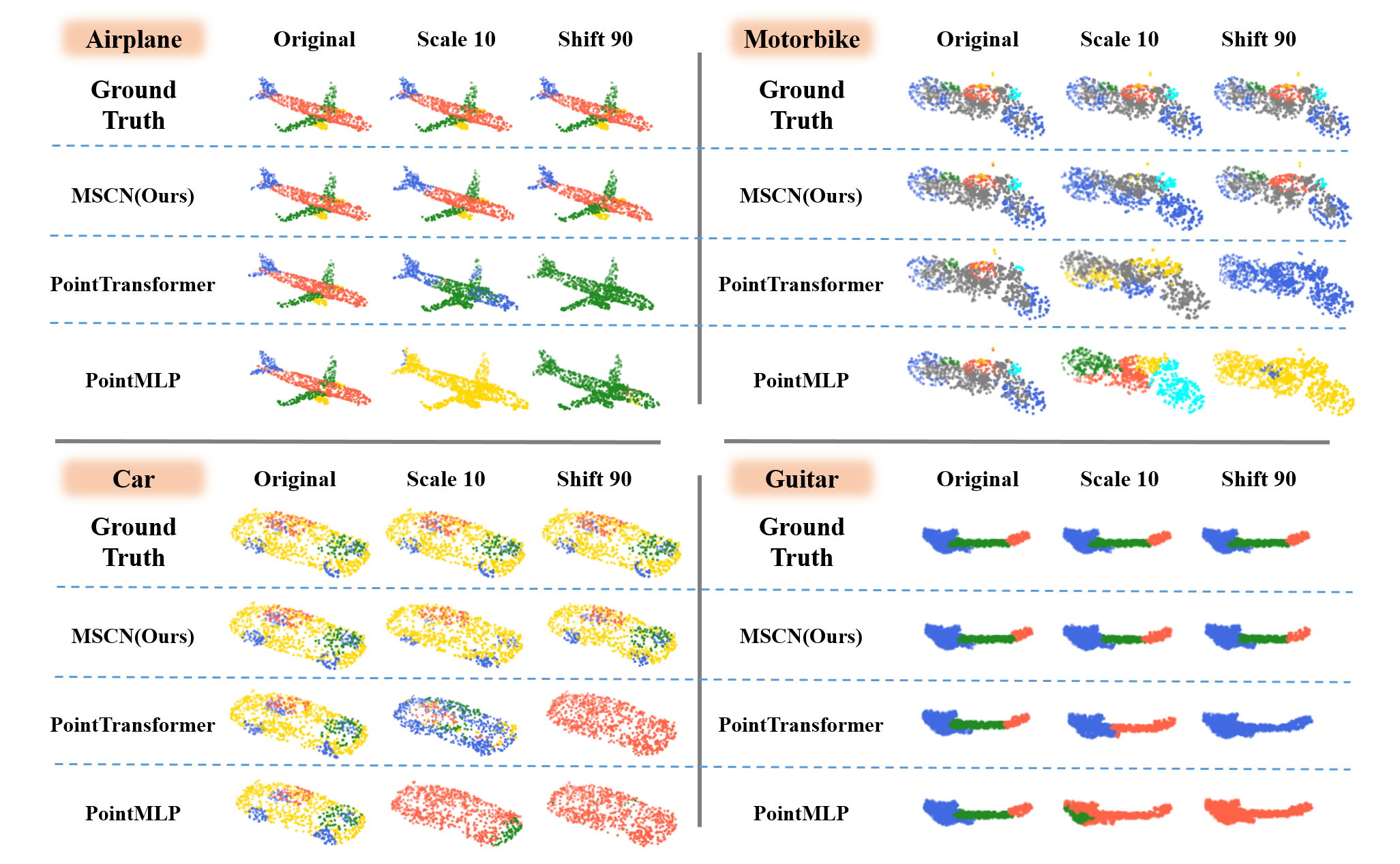}}
    \caption{Examples of geometric transformation on ShapeNetPart~\cite{ShapeNet}, where \textbf{scale 10} represents a transformation where the original point cloud is scaled by a factor of 10, while \textbf{shift 90} indicates a random translation of the point cloud by a distance of 90 units.}
    \label{ShapeNet} 
\end{figure*}

We first conducted experiments on the ShapeNetPart~\cite{ShapeNet} dataset for part segmentation. Table~\ref{table:ShapeNet} summarizes the part segmentation results, where mIoU values are reported for each method. While SOTA models exhibit high performance on the given dataset, MSCN's performance is slightly inferior compared to SOTA models. This performance gap can be explained by the inherent trade-off between robustness and accuracy, as discussed in \cite{Acc-Robust-trade-off}. Our MSCN were designed to generalize well to harsh or out-of-distribution scenarios while preserving accuracy on natural data. However, Zhang et.al~\cite{Acc-Robust-trade-off} showed this approach often leads to a shift in focus from minimizing standard classification errors (natural error) to improving robustness, which can degrade accuracy in ideal.

\begin{figure}[t]
\includegraphics[width=0.95\columnwidth]{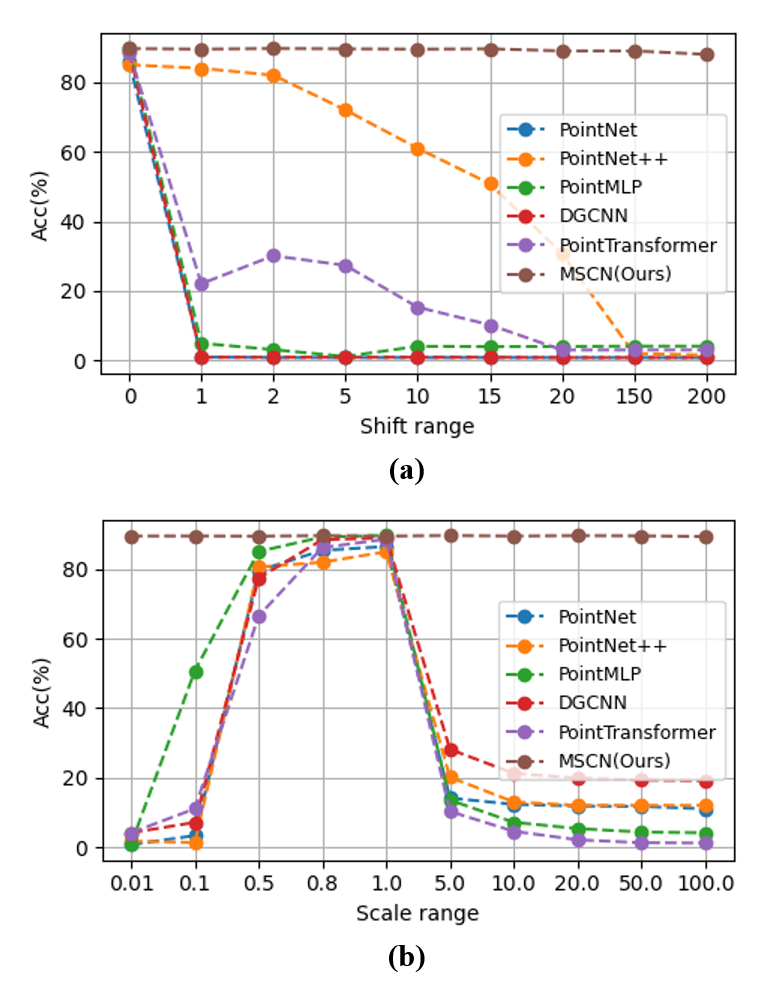}
\centering \vspace{-0.5cm}
\caption{Evaluation of Geometrical Invariance Properties on ModelNet40~\cite{ModelNet40}. Each graph illustrates the accuracy of various models under specific geometric transformations applied to point clouds. (a) Translation: Accuracy as the translation magnitude increases sequentially from 0 to 200 units. (b) Scaling: Accuracy as the scale factor increases sequentially from 0.01 to 100.}
\label{ModelNet}
\end{figure}

To further evaluate robustness of MSCN, we conducted experiments to visualize the robustness of MSCN compared to PointMLP and PointTransformer under geometric transformations. Figure~\ref{ShapeNet} illustrates the results of scaling and translating point clouds. MSCN demonstrates robust part segmentation even under significant geometrical transformations, whereas PointMLP and PointTransformer show sensitivity to these changes, resulting in suboptimal performance. These results emphasize MSCN’s ability to generalize effectively across various geometric variations.

\begin{table}[!t]
\caption{Classification results on ModelNet40~\cite{ModelNet40}.}
\centering
\begin{tabular}{c|c|c}
\hline
\textbf{Method} & \textbf{\# points} & \textbf{Acc. (\%)} \\ \hline
     PointNet++~\cite{pointnet+}       & 1024        & 93.8 \\ 
     MVTN~\cite{MVTN} &  1024 & 93.8\\ 
     RepSurf-U~\cite{RepSurf} &   1024  & 94.4 \\ 
     PointTransformer V2~\cite{transformerV2}    & 1024  &  94.2  \\ 
     GPSFormer~\cite{GPSFormer} & 1024 & 94.2 \\
     PointNet~\cite{pointnet}      & 1024       & 89.2 \\ 
     PointNeXt~\cite{pointNext}      & 1024       & 94.0 \\ 
     DGCNN~\cite{dgcnn}       & 1024       & 92.9 \\ 
     GCN~\cite{gcn}       & 1024       & 92.1 \\ 
     PointMamba~\cite{pointmamba}       &   1024    & 93.6 \\ 
     PointMLP~\cite{pointMLP}        &  1024   & \textbf{94.5} \\
     PointTransformer~\cite{transformer}       &   1024    & 93.7 \\
     \hline
     MSCN(Ours)   & 1024      &  92.2 \\ \hline
\end{tabular}
\label{table:ModelNet}
\end{table}

\begin{table*}[htbp]
\caption{Classification results on the ScanObjectNN~\cite{ScanObjectNN} dataset. This table summarizes the performance of methods under different experimental settings, including self-supervised pretraining, supervised learning, and the evaluation of MSCN's translation and scale invariance. PT, RT, and RS indicates Pretrained models, Random Translation, and Random Scaling, respectively.}
\centering
\begin{tabular}{lcccccccc}
\toprule
\textbf{Method} & \textbf{PT} & \textbf{RT} & \textbf{RS} & \textbf{OBJ\_BG $\uparrow$} & \textbf{OBJ\_ONLY $\uparrow$} & \textbf{PB\_T50\_RS $\uparrow$} & \\
\midrule

\multicolumn{8}{c}{\textit{With Self-supervised Pre-training}} \\
Transformer~\cite{transformer} & OcCo~\cite{OcCo} &  $\times$ &  $\times$ & 84.85 & 85.54 & 78.79  \\
Point-BERT~\cite{Point-BERT} & IDPT~\cite{IDPT} &  $\times$ &  $\times$ & 88.90 & 88.10 & 85.01  \\
MaskPoint~\cite{MaskPoint}  & MaskPoint~\cite{MaskPoint} &  $\times$ &  $\times$ & 89.10 & \textbf{88.20} & 85.02  \\
PointMamba~\cite{pointmamba} & IDPT~\cite{IDPT} &  $\times$ &  $\times$ & 89.10 & 88.10 & 85.03 \\
\midrule

\multicolumn{8}{c}{\textit{Supervised Learning Only}} \\
PointNet~\cite{pointnet} & $\times$ &  $\times$ &  $\times$ & 73.3 & 79.2 & 68.0  \\
PointNet++~\cite{pointnet+} & $\times$ &  $\times$ &  $\times$ & 82.3 & 84.3 & 75.4  \\
DGCNN~\cite{dgcnn} & $\times$ &  $\times$ &  $\times$ & 82.6 & 86.2 & 78.1 \\
PointCNN~\cite{PointCNN} & $\times$ &  $\times$ &  $\times$ & 86.1 & 85.5 & 80.5  \\
PointNeXt~\cite{pointNext} & $\times$ &  $\times$ &  $\times$ & - & -& \textbf{87.7} $\pm$0.4  \\

PCT~\cite{PCT}  & $\times$ &  $\times$ &  $\times$ & - & - & 82.48 \\
PointMamba~\cite{pointmamba} & $\times$ &  $\times$ &  $\times$ & 88.30 & 87.78 & 82.48  \\
PointMLP~\cite{pointMLP} & $\times$ &  $\times$ &  $\times$ & - &- & 85.4 $\pm$0.3  \\
PointTransformer~\cite{transformer} & $\times$ &  $\times$ &  $\times$ & 79.86 & 80.55 & 77.24  \\
MSCN(Ours)  & $\times$ &  $\times$ &  $\times$ & \textbf{89.33}& 86.91 & 83.03  \\
\midrule
\multicolumn{8}{c}{\textit{Translation and Scale Invariance of MSCN}} \\
MSCN(Ours)  & $\times$ &   \checkmark &  $\times$ & 89.67& 86.30 & 83.83  \\
MSCN(Ours)  & $\times$ &  $\times$ &   \checkmark & 89.67& 86.30 & 83.83 \\
MSCN(Ours)  & $\times$ & \checkmark &   \checkmark & 89.66 & 86.30 & 83.83 \\
\bottomrule
\end{tabular}
\label{ScanObjectNN}
\end{table*}

Furthermore, we conducted experiments on the ModelNet40~\cite{ModelNet40} dataset for object classification. Table~\ref{table:ModelNet} presents the classification results, where MSCN is slightly lower than some SOTA models. Although SOTA models demonstrate strong performance on this dataset, their sensitivity to geometric transformations can result in significant performance degradation in more challenging scenarios.
To further investigate this, Figure~\ref{ModelNet} illustrates the performance of models under geometric transformations such as translation and scaling. MSCN consistently exhibits robust performance across all transformations, maintaining stable accuracy despite severe geometric changes. In contrast, other methods experience substantial drops in accuracy, highlighting their vulnerability to such transformations.
These results underscore the strength of MSCN in terms of geometrical invariance, demonstrating its potential for real-world applications where robustness to geometric variations is critical.

However, it is important to note that ModelNet40 and ShapeNetPart are derived from CAD models, which are generated in highly idealized environments. In contrast, real-world point clouds are often sparse and only capture surface-level details, which makes them fundamentally different from synthetic datasets. To address this disparity, we extended our experiments to real-world point clouds, enabling a more comprehensive evaluation of our MSCN's performance in practical scenarios.

\subsection{Evaluation on Indoor Real-World Datasets: Comparable Performance and Robustness}

We conducted experiments on the ScanObjectNN~\cite{ScanObjectNN} dataset to evaluate the performance of MSCN on real-world indoor point clouds. Table~\ref{ScanObjectNN} summarizes the classification results, demonstrating that MSCN achieves performance comparable to the SOTA models in standard evaluation settings. Furthermore, MSCN exhibits remarkable robustness under geometric transformations such as scaling and translation, maintaining stable performance and demonstrating its strength in handling these variations.

However, it is important to note that ScanObjectNN consists of point clouds captured in indoor environments, which are relatively dense and structured. In contrast, point clouds captured in outdoor environments, such as those used in autonomous driving, are significantly sparser and often acquired under harsh conditions, including occlusions and uneven sampling. To address these challenges, we extended our experiments to self-driving vehicle datasets, enabling a more comprehensive evaluation of MSCN's robustness in real-world scenarios.

\begin{table}[t]
\caption{Intra-domain classification results on KITTI~\cite{KITTI}, nuScenes~\cite{nuScene}, and PanKyo~\cite{PanKyo} datasets. The bold \textbf{black}, \textcolor{blue}{\textbf{blue}} indicate the first and the second best performance on each scenario, respectively.}
  \centering
  {\small{
    \begin{tabular}{c||c|c }
    \hline
    Train/Test Dataset   &   Method & Acc.(\%)\\
    \hline\hline
&PointNet~\cite{pointnet} &  99.5\\
&PointNeXt~\cite{pointNext} &99.6\\
&PointMLP~\cite{pointMLP} & \textcolor{blue}{\textbf{99.9}} \\
KITTI&DGCNN~\cite{dgcnn} &99.6\\
(German, 64CH)&GCN~\cite{gcn} & 96.3 \\
&SCN~\cite{PanKyo} & 99.2\\
&PointTransformer~\cite{transformer}& 99.3\\
&PointMamba~\cite{pointmamba}& \textbf{100}\\
&MSCN(ours) & 97.5 \\ 
\hline

&PointNet~\cite{pointnet} &  92.6\\
&PointNeXt~\cite{pointNext} & 93.6\\
&PointMLP~\cite{pointMLP} & \textcolor{blue}{\textbf{96.6}} \\
Pankyo&DGCNN~\cite{dgcnn} &95.5\\
(Korea, 128CH)&GCN~\cite{gcn} & 93.4 \\
&SCN~\cite{PanKyo} & 93.0\\
&PointTransformer~\cite{transformer}& \textbf{98.5}\\
&PointMamba~\cite{pointmamba}& 95.5\\
&MSCN(ours) & 94.5 \\ 
\hline

&PointNet~\cite{pointnet} &  95.6\\
&PointNeXt~\cite{pointNext} & 95.5\\
&PointMLP~\cite{pointMLP} & 94.7 \\
\ygk{nuScenes} & DGCNN~\cite{dgcnn} &95.5 \\
(U.S., 32CH)&GCN~\cite{gcn} & 87.6 \\
&SCN~\cite{PanKyo} & 90.8\\
&PointTransformer~\cite{transformer}& \textcolor{blue}{\textbf{96.2}}\\
&PointMamba~\cite{pointmamba}& \textbf{97.9}\\
&MSCN(ours) & 92.0 \\ 
\hline
    \end{tabular} 
  }}
  \label{table1}
\end{table}

\subsection{Evaluation on Self-Driving Datasets: Cross-Domain Robustness}
Our methods focus on maintaining consistent object recognition performance regardless of the type of point cloud data encountered. This means they are designed to work well not only with the data they were originally trained on but also with data from different sources. To thoroughly assess their effectiveness, we conducted two types of evaluations. Both intra-domain and cross-domain performances were evaluated to provide a comprehensive understanding of the capabilities of the methods and to ensure their reliability and effectiveness across various datasets.

First,  the performances are demonstrated within the KITTI~\cite{KITTI}, nuScenes~\cite{nuScene}, and PanKyo~\cite{PanKyo} datasets to capture classification performance seamlessly using the training dataset on the test dataset. \ygk{Table~\ref{table1}} shows the classification performance within a single domain. These results measure the model's ability to classify point clouds within the same domain used for training. While MSCN achieves competitive results, it lags slightly behind the SOTA methods in terms of accuracy. For example, the average intra-domain accuracy of MSCN across all datasets is 94.7\%, compared to PointTransformer's 98\%. This performance gap can be explained from the bias-variance trade-off~\cite{ascn, bias-variance}. MSCN, designed for robust domain-invariant feature extraction, prioritizes reducing model variance to generalize effectively across diverse domains. As a result, its intra-domain performance exhibits slightly higher bias, which is a common trade-off observed in models designed with high generalization capacity. Specifically, while other SOTA models may minimize bias within the training domain, they often exhibit poor performance in domain-shift scenarios due to higher variance.

To demonstrate MSCN's robustness, we further evaluated its performance under cross-domain scenarios as illustrated in Table~\ref{table2}, where substantial domain shifts between training and testing datasets. PointNet, PointNeXt, PointMLP, DGCNN, PointTransformer, and PointMamba demonstrate good performance when the domain shifts from simulation to the real world (Sim2Real) but show significantly lower performance when shifting from one real-world dataset to another (Real2Real). On the other hand, GCN and SCN show robust performance in Real2Real domain change scenarios but are sensitive in Sim2Real scenarios.

However, our MSCN demonstrates robust performance in both Real2Real and Sim2Real scenarios compared to the aforementioned methods. In particular, MSCN surpasses all SOTA methods with an average cross-domain accuracy of 82\%, significantly higher than PointTransformer's 66.2\%. This 15.8\% improvement highlights the effectiveness of MSCN in handling domain shifts, which is critical for real-world applications such as autonomous driving and robotics.

The results collectively demonstrate that while MSCN’s intra-domain performance is slightly lower due to the bias-variance trade-off, its superior generalization ability in cross-domain scenarios validates its design focus on robust domain-invariant feature extraction.

\begin{table*}[!t]
\caption{Cross-domain classification results on KITTI~\cite{KITTI}, nuScenes~\cite{nuScene}, and PanKyo~\cite{PanKyo} datasets. The bold \textbf{black}, \textcolor{blue}{\textbf{blue}} indicate the first and the second best performance on each scenario, respectively.} 
\resizebox{\textwidth}{!}{%
\begin{tabular}{c|c|cc|cc|cc|ccc|c}
\hline
 & &
  \multicolumn{2}{c|}{KITTI (64CH)} &
  \multicolumn{2}{c|}{PanKyo (128CH)} &
  \multicolumn{2}{c|}{nuScenes (32CH)} &
  \multicolumn{3}{c|}{Simulation (32CH)} &
  \multicolumn{1}{l}{Average} \\ \cline{3-11}
\multirow{-2}{*}{Methods} & \multirow{-2}{*}{Param. (M)} &
  PanKyo &
  nuScenes &
  KITTI &
  nuScenes &
  KITTI &
  PanKyo &
  KITTI &
  \multicolumn{1}{l}{PanKyo} &
  \multicolumn{1}{l|}{nuScenes} &
  \multicolumn{1}{l}{Acc(\%)} \\ \hline
PointNet~\cite{pointnet} & 3.0&
  59.3 &
  62.8 &
  6.2 &
  21.8 &
  36.3 &
  47.9 &
  82.6 &
  78.5 &
  71.8 &
  51.9 \\
PointNeXt~\cite{pointNext}&14.1&
  {\color[HTML]{000000} \textcolor{blue}{\textbf{67.6}}} &
  31.4 &
  5.8 &
  17.3 &
  31.6 &
  18.6 &
  82.7 &
 \textcolor{blue}{\textbf{88.6}} &
 76.1 &
  46.6 \\
PointMLP~\cite{pointMLP}& 13.2 &
  17.7 &
  80.4 &
  {\color[HTML]{000000} \textcolor{blue}{\textbf{87.7}}} &
  64.0 &
  47.1 &
  15.2 &
  81.0 &
  84.6 &
  35.2 &
  57.0 \\
DGCNN~\cite{dgcnn}& 4.2&
  14.5 &
  38.0 &
  6.2 &
  25.6 &
  72.2 &
  20.1 &
  81.2 &
  82.7 &
  \textbf{94.6} &
  48.3 \\
GCN~\cite{gcn}& 1.2&
  60.9 &
  60.6 &
  83.0 &
  64.1 &
  \textbf{92.8}&
  \textbf{86.2}&
  20.7 &
  16.7 &
  23.8 &
  56.5 \\
SCN~\cite{PanKyo}& 3.5&
  42.6 &
  72.3 &
  84.4 &
  {\color[HTML]{000000} \textcolor{blue}{\textbf{67.3}}} &
  {\color[HTML]{000000} \textcolor{blue}{\textbf{92.1}}} &
  74.3 &
  30.9 &
  48.4 &
  29.1 &
 60.2 \\
PointTransformer~\cite{transformer}&9.6 &
7.9 &  \textcolor{blue}{\textbf{88.1}} &  77.3 & 73.3 & 79.1 & 8.8  & 80.5 & 87.1 & \textcolor{blue}{\textbf{94.0}} & \textcolor{blue}{\textbf{66.2}}  \\
PointMamba~\cite{pointmamba} & 12.3
& 6.6 & \textbf{90.1}& 11.5 & 37.5 & 61.6 & 30.6 & \textbf{96.9} & \textbf{90.9} & 89.3& 57.2 \\
\rowcolor[HTML]{E0E0E0} 
MSCN(ours) & 4.2&
  \textbf{81.7} &
  75.3 &
  \textbf{88.2} &
  \textbf{78.5} &
  \textbf{92.8} &
  \textcolor{blue}{\textbf{84.8}} &
  \textcolor{blue}{\textbf{83.5}} &
  83.9 &
  69.2 &
  \textbf{82.0} \\ 
  \hline
\end{tabular}
}
\label{table2}
\end{table*}

\begin{table*}[!t]
\caption{Evaluation on Inference time across AV datasets. All values are measured in milliseconds (ms).} 
\resizebox{\textwidth}{!}{%
\begin{tabular}{c|c|ccc|ccc|ccc|ccc|c}
\hline
 & &
  \multicolumn{3}{c|}{KITTI} &
  \multicolumn{3}{c|}{PanKyo} &
  \multicolumn{3}{c|}{nuScenes} &
  \multicolumn{3}{c|}{Simulation} &
  \multicolumn{1}{l}{Average} \\ \cline{3-14}
\multirow{-2}{*}{Methods} & \multirow{-2}{*}{Param.(M)} &
  KITTI & PanKyo & nuScenes &
  KITTI & PanKyo & nuScenes &
  KITTI & PanKyo & nuScenes &
  KITTI & PanKyo & nuScenes &  
  Time(ms) \\ \hline
  
PointNet~\cite{pointnet} & 3.0
 & 4.1& 2.0& 0.8& 3.9& 1.9& 0.7& 3.9& 2.2& 0.8 & 4.0& 2.1& 0.8 &  2.3  \\
 
PointNeXt~\cite{pointNext} & 14.1
 & 38.2& 37.5& 37.3& 38.2& 37.4& 37.1 & 38.1& 37.5& 37.2  & 38.1& 37.4& 37.2 &37.6 \\
 
PointMLP~\cite{pointMLP} & 13.2
 & 32.4& 31.1& 31.0 & 32.4& 31.1& 31.0 & 32.6& 31.4& 31.1 &  32.5& 31.2& 31.1&  31.6 \\
 
DGCNN~\cite{dgcnn} & 4.2
 & 10.0 & 8.8& 8.2& 10.0 & 8.8& 8.2& 10.0 &8.8 & 8.1  & 10.1& 8.9& 8.2 & 9.0  \\
 
GCN~\cite{gcn} & 1.2
 & 3.6& 2.9& 2.6& 3.6& 3.0& 2.6& 3.6& 2.9& 2.6 & 3.6& 2.8& 2.5 &  3.0 \\
 
SCN~\cite{PanKyo} & 3.5

 & 4.0& 3.2& 2.8& 4.2& 3.4& 2.8& 4.5& 3.3& 2.9& 4.1& 3.1& 2.7&  3.4  \\
PointTransformer~\cite{transformer} & 9.6

 & 36.5& 34.6& 34.1& 35.8 & 36.3 & 35.3 &36.5& 34.6& 34.1 & 36.4& 34.5& 34.0 & 35.2  \\

PointMamba~\cite{pointmamba}  & 12.3
 & 5.4& 4.1& 3.4& 5.4&  4.1& 3.4& 5.5& 4.1& 3.4 & 5.4& 4.1& 3.4 &  4.3 \\

\rowcolor[HTML]{E0E0E0} 
MSCN(ours) & 4.2
 & 4.7& 3.6& 3.1& 4.9& 3.6& 3.2& 4.4& 3.6& 3.2& 4.7& 3.6& 3.1 &  3.8 \\

  \hline
\end{tabular}
}
\label{table3}
\end{table*}

\begin{figure*}[!t]
    \centerline{\includegraphics[width=0.95\textwidth]{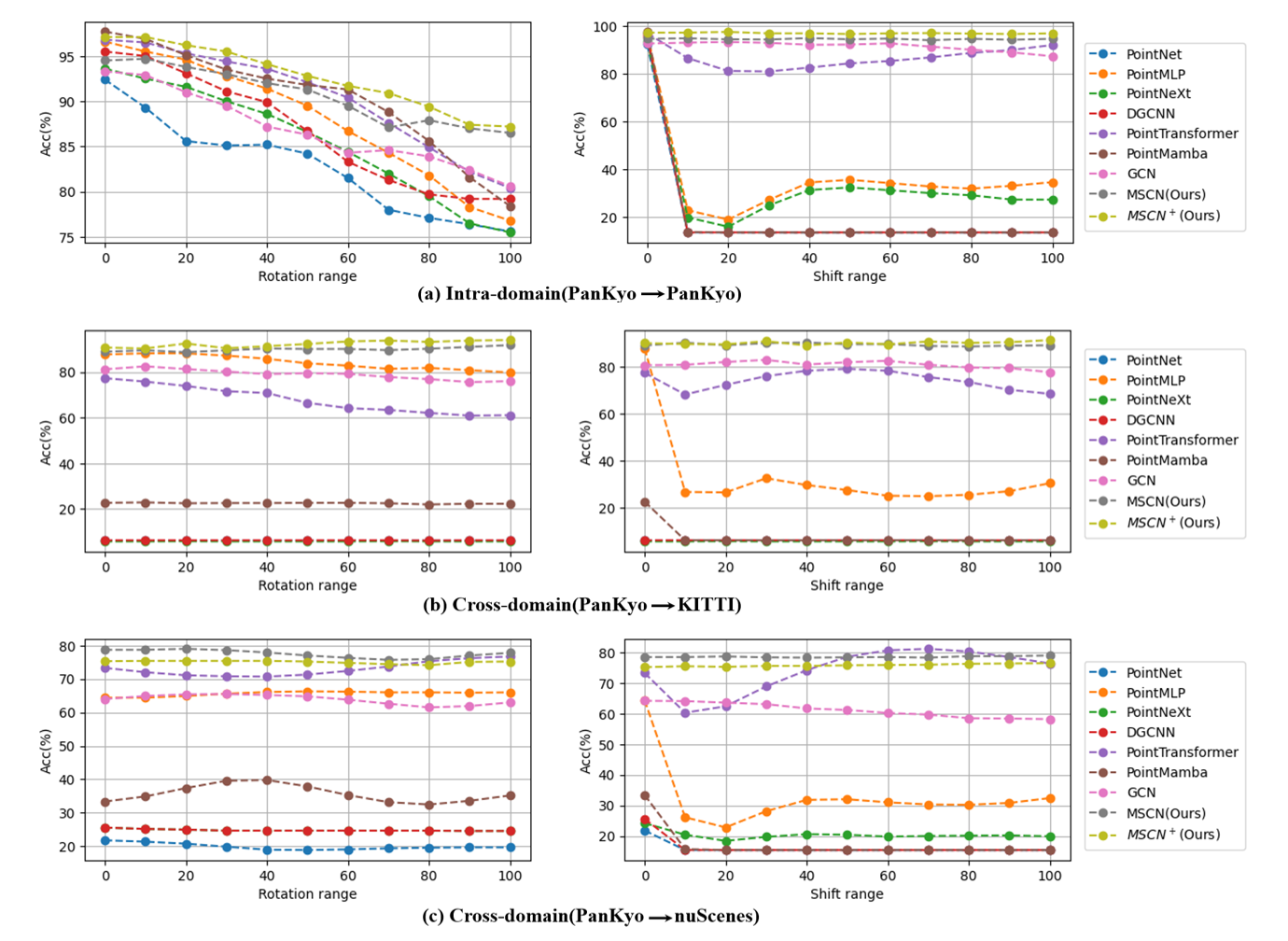}}
    \caption{Evaluation of Invariance Properties. All models were trained on the PanKyo \cite{PanKyo} dataset and then tested on (a) PanKyo\cite{PanKyo}, (b) KITTI\cite{KITTI}, and (c) nuScenes\cite{nuScene} datasets. The graphs on the left illustrate the performance of the models when point clouds were rotated along the upward direction. The graphs on the right show the results when the models were tested with randomly shifted point clouds within a certain distance in all directions.}
    \label{Fig6} 
\end{figure*}

Furthermore, Table~\ref{table3} presents the inference time required to predict the class of a single point cloud in both intra-domain and cross-domain settings. Models such as PointMLP, PointNeXt, and PointTransformer exhibit higher inference times due to their significantly large number of parameters. In contrast, although PointMamba is also Transformer-based and has a large parameter count, its inference time is relatively low. This efficiency can be attributed to its design, which likely focuses on reducing FLOPs (Floating Point Operations Per Second) to enhance computational efficiency.

In particular, the average inference time of MSCN is only 3.8 ms, which is well suited for real-world applications such as autonomous driving. Modern rotary LiDAR sensors typically operate at speeds of 10 Hz to 20 Hz, meaning one full rotation takes 0.05 to 0.1 seconds. Given that MSCN can process a single point cloud within 3.8 ms, it is capable of handling multiple scans within a single rotation, ensuring real-time performance without bottlenecking the data pipeline. This efficient processing time, combined with robust cross-domain performance, demonstrates the viability of MSCN for deployment in real-world autonomous systems.

\begin{table}[t]
\caption{Layer configuration evaluation results on KITTI~\cite{KITTI}, nuScenes~\cite{nuScene}, PanKyo~\cite{PanKyo}, and simulation~\cite{simulation} datasets.} 
\begin{center} 
\begin{tabular}{c|c|c|c}
\hline
Training Dataset & Test Dataset   &   Layer Configuration & Acc.(\%)\\
\hline\hline

\multirow{4}{*}{KITTI}& \multirow{2}{*}{PanKyo} & only SCL & 78.9 \\
& & MSCN(SCL+SAL) & \textbf{81.7} \\
 \cline{2-4}
&\multirow{2}{*}{nuScenes} & only SCL & 74.8 \\
& & MSCN(SCL+SAL) & \textbf{75.3} \\
 \cline{2-4}
  \hline

\multirow{4}{*}{PanKyo}& \multirow{2}{*}{KITTI} & only SCL & 84.9 \\
& & MSCN(SCL+SAL)& \textbf{88.2} \\
 \cline{2-4}
&\multirow{2}{*}{nuScenes} & only SCL & 71.6 \\
& & MSCN(SCL+SAL) & \textbf{78.5} \\
 \cline{2-4}
  \hline

\multirow{4}{*}{nuScenes}& \multirow{2}{*}{KITTI} & only SCL & 92.1 \\
& & MSCN(SCL+SAL)& \textbf{92.8} \\
 \cline{2-4}
&\multirow{2}{*}{PanKyo} & only SCL & 78.4 \\
& & MSCN(SCL+SAL) & \textbf{84.8} \\
 \cline{2-4}
  \hline

& \multirow{2}{*}{KITTI} & only SCL & 71.3 \\
& & MSCN(SCL+SAL)& \textbf{83.5} \\
 \cline{2-4}
\multirow{2}{*}{Simulation}& \multirow{2}{*}{PanKyo} & only SCL & 75.1 \\
& & MSCN(SCL+SAL) & \textbf{83.9} \\
 \cline{2-4}
&\multirow{2}{*}{nuScenes} & only SCL & 45.9 \\
& & MSCN(SCL+SAL) & \textbf{69.2} \\
 \cline{2-4}
  \hline
\end{tabular}
\end{center}

\label{table4}

\end{table}

\begin{table}[t]
\caption{Unseen domain generation effects on KITTI~\cite{KITTI}, nuScenes~\cite{nuScene}, PanKyo~\cite{PanKyo}, and simulation~\cite{simulation} datasets. The $\dag$ mark indicates the method trained with our unseen domain generation method, achieved from equation 13 and equation 19.} 
\begin{center} 
\begin{tabular}{c|c|c|c|c}

\hline
Training Dataset & Test Dataset   &   Method & Acc.(\%) & Time (ms)\\
\hline\hline

& \multirow{2}{*}{KITTI} & MSCN & 97.5 & 4.7 \\
& & MSCN$^{\dag}$& \textbf{99.7} & 4.4 \\
 \cline{2-5}
\multirow{2}{*}{KITTI}& \multirow{2}{*}{PanKyo} & MSCN & \textbf{81.7} & 3.6 \\
& & MSCN$^{\dag}$ & \textbf{81.7} & 3.5\\
 \cline{2-5}
&\multirow{2}{*}{nuScenes} & MSCN & 75.3 & 3.1\\
& & MSCN$^{\dag}$ & \textbf{82.7} &3.1\\
 \cline{2-5}
  \hline

& \multirow{2}{*}{KITTI} & MSCN & 88.2&4.9 \\
& & MSCN$^{\dag}$& \textbf{89.8} &4.4\\
 \cline{2-5}
\multirow{2}{*}{Pankyo}& \multirow{2}{*}{PanKyo} & MSCN & 94.5 &3.6\\
& & MSCN$^{\dag}$ & \textbf{98.3} &3.5\\
 \cline{2-5}
&\multirow{2}{*}{nuScenes} & MSCN & \textbf{78.5} &3.2\\
& & MSCN$^{\dag}$ & 75.3 &3.1\\
 \cline{2-5}
  \hline

& \multirow{2}{*}{KITTI} & MSCN & 92.8 &4.4\\
& & MSCN$^{\dag}$& \textbf{95.4} & 4.2\\
 \cline{2-5}
\multirow{2}{*}{nuScenes}& \multirow{2}{*}{PanKyo} & MSCN & 84.8 &3.6\\
& & MSCN$^{\dag}$ & \textbf{86.6} &3.6\\
 \cline{2-5}
&\multirow{2}{*}{nuScenes} & MSCN & 92.0 &3.2\\
& & MSCN$^{\dag}$ & \textbf{94.6} &3.0\\
 \cline{2-5}
  \hline

& \multirow{2}{*}{KITTI} & MSCN & 83.5&4.7 \\
& & MSCN$^{\dag}$& \textbf{95.7} &4.4\\
 \cline{2-5}
\multirow{2}{*}{Simulation}& \multirow{2}{*}{PanKyo} & MSCN & 83.9 &3.6\\
& & MSCN$^{\dag}$ & \textbf{96.2} &3.5\\
 \cline{2-5}
&\multirow{2}{*}{nuScenes} & MSCN & 69.2 &3.1\\
& & MSCN$^{\dag}$ & \textbf{87.1} &3.1\\
 \cline{2-5}
  \hline
\end{tabular}
\end{center}
\label{table5}

\end{table}

\begin{table}[t]
\caption{\ygk{Loss ablation results of MSCN$^{\dag}$ trained on KITTI dataset and evaluated across KITTI~\cite{KITTI}, nuScenes~\cite{nuScene}, and PanKyo~\cite{PanKyo}.}} 
\begin{center} 
\begin{tabular}{c|c|c|c}

\hline
Training Dataset & Test Dataset  &   Method & Acc.(\%)\\
\hline\hline

\multirow{15}{*}{KITTI} & \multirow{5}{*}{KITTI} & w/o $L_{recon}$ & \textbf{99.7} \\
& & w/o $L_{div}$ & 99.6 \\
& & w/o $L_{adv}$ & 99.4 \\
& & w/o $L_{gen, ce}$ & \textbf{99.7} \\
\cline{3-4}
& & MSCN$^{\dag}$& \textbf{99.7} \\

 \cline{2-4}

 & \multirow{5}{*}{PanKyo} & w/o $L_{recon}$ & 78.4 \\
& & w/o $L_{div}$ & 71.9 \\
& & w/o $L_{adv}$ & 59.8 \\
& & w/o $L_{gen, ce}$ & 63.2 \\
\cline{3-4}
& & MSCN$^{\dag}$& \textbf{81.7} \\

 \cline{2-4}

& \multirow{5}{*}{nuScenes} & w/o $L_{recon}$ & 82.1 \\
& & w/o $L_{div}$ & 80.4 \\
& & w/o $L_{adv}$ & 79.0 \\
& & w/o $L_{gen, ce}$ & 81.6 \\
\cline{3-4}
& & MSCN$^{\dag}$& \textbf{82.7} \\

\hline
  
\end{tabular}
\end{center}
\label{table6}

\end{table}

Finally, to further reflect the operational conditions of real-world AVs, we evaluate the robustness of our model not only under sensor configuration changes but also under continuous geometric variations in point clouds that naturally occur during driving. As AVs move, the surrounding point clouds are constantly subjected to shifts and rotations due to changes in viewpoint and vehicle trajectory. To replicate this, we progressively rotate the point clouds from 0$^{\circ}$ to 100$^{\circ}$ in 10$^{\circ}$ increments and randomly shift them within a range of 0 to 100 units, while simultaneously considering cross-domain sensor configuration changes.

Figure~\ref{Fig6} presents six different scenarios that combine rotation and shift transformations in both intra-domain and cross-domain settings. In the intra-domain experiments, most baseline models suffer a sharp decline in accuracy as the rotation or shift magnitude increases, whereas our MSCN and MSCN$^\dag$ show only a gradual decrease, demonstrating strong resilience to geometric perturbations. More importantly, in cross-domain scenarios such as PanKyo $\rightarrow$ KITTI and PanKyo $\rightarrow$ nuScenes, our models consistently maintain superior accuracy under substantial rotations and shifts, outperforming all baselines. These results confirm that MSCN and MSCN$^\dag$ are not only effective at handling domain shifts across heterogeneous LiDAR sensors, but are also robust to dynamic geometric transformations, making them well-suited for deployment in the challenging and continuously changing environments encountered by real-world AVs.

\subsection{Ablation study}

We evaluate the compatibility of layer configurations of MSCN, the performance enhancement when MSCN was trained with synthetic point cloud datasets from our unseen domain generation method, \ygk{and the impact of each loss function used for domain expansion}.

First, to assess the compatibility of layer configurations of MSCN, we compare the cases where there are configurations of both SCL and SAL, and only SCL. The former consists of four SCL and two SAL, which is illustrated in Figure~\ref{Fig2} (a). The latter is composed of five SCL that are SCL(3, 32), SCL(32, 64), SCL(64, 128), SCL(128, 256), and SCL(256, 1024). Table~\ref{table4} shows the performance of the methods in layer configurations where domain shift scenarios. MSCN outperforms the case that has only SCL without SAL during all domain shift scenarios, with an average performance that is 7.2\% higher than the case that only consist of SCL.

To evaluate the impact of our unseen domain generalization methods on improving the leaning of domain-invariant feature representation, we performed various experiments based on both intra-domain and cross-domain classification, which are illustrated in Table~\ref{table5}. MSCN$^{\dag}$, which can increase the amount of training without overfitting, performs better than MSCN for single-domain classification. Note that MSCN does not include a domain generation module. With the domain generation module, MSCN$^{\dag}$ demonstrates even more robust performance than MSCN in various domain change scenarios, with an average improvement of 5.8\%, indicating that our proposed unseen domain generation method ensures domain generalization ability.

\ygk{Finally, to evaluate the contribution of each loss component, we conducted a loss ablation study using MSCN$^{\dag}$ trained on the KITTI dataset and evaluated across KITTI, PanKyo, and nuScenes, illustrated in Table~\ref{table6}. 
In the intra-domain setting (KITTI$\rightarrow$KITTI), all variants achieve comparable accuracy ($>$99\%), since the data distribution between training and testing domains is consistent and the model can easily fit the geometry of a single domain. 
In contrast, the cross-domain results (KITTI$\rightarrow$PanKyo and KITTI$\rightarrow$nuScenes) show clear performance degradation when any loss term is removed, demonstrating the complementary roles of each component. 
Specifically, removing the adversarial loss ($L_{\text{adv}}$) severely reduces accuracy, confirming its importance in aligning feature distributions across domains. 
The generator classification loss ($L_{\text{gen,ce}}$) ensures semantic consistency between generated and real features, and its absence leads to unstable domain alignment. 
The reconstruction loss ($L_{\text{recon}}$) preserves geometric structure and mitigates noise amplification, while the diversity loss ($L_{\text{div}}$) encourages the generation of varied unseen domains, preventing mode collapse. 
Without these loss terms, the model tends to overfit on the source domain due to the large number of training epochs, resulting in degraded cross-domain generalization.
These findings verify that each loss contributes synergistically to achieving robust and domain-invariant feature representation.}

\section{Conclusion}
\label{sec:conclusion}

In this paper, we introduced MSCN, which learns domain-invariant feature representations by capturing the geometric information of 3D point clouds from the standpoint of both local context and overall context. \ygk{Furthermore, we developed an unseen-domain expansion framework that enables MSCN to learn more robust and transferable features across different sensing conditions. Comprehensive cross-domain evaluations demonstrated that our MSCN achieved an average accuracy of \textbf{82.0\%}, surpassing the strong baseline PointTransformer by \textbf{15.8\%}, while MSCN$^{\dag}$ further improved performance by an additional \textbf{5.8\%}. These quantitative results confirm the effectiveness of our approach in addressing sensor-dependent and domain-shift challenges in point cloud classification for autonomous vehicles. } In light of these findings, our future research endeavors will focus on expanding our method to object detection tasks for real-world applications that require domain-invariant recognition, such as autonomous driving, robotics, and virtual reality. 

\label{section6}

\section*{Acknowledgment}
This work was supported by the Korea National Police Agency(KNPA) under the project `Development of the autonomous driving patrol service for active prevention and response to traffic accidents' (RS-2024-00403630) and the BK21 FOUR program of the National Research Foundation Korea funded by the Ministry of Education(NRF5199991014091).

\begin{IEEEbiography}[{\includegraphics[width=1in,height=1.25in,clip,keepaspectratio]{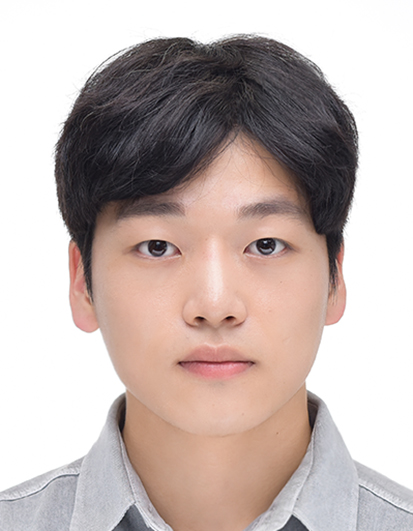}}]{Younggun Kim} received B.S. degree in mechanical engineering from Ajou University, Suwon, South Korea, in 2024. He is currently pursuing the M.S. degree in civil engineering with University of Central Florida, Florida, USA. His research interests include artificial intelligence, machine learning, and transportation.
\end{IEEEbiography}

\vspace{-1cm}

\begin{IEEEbiography}[{\includegraphics[width=1in,height=1.25in,clip,keepaspectratio]{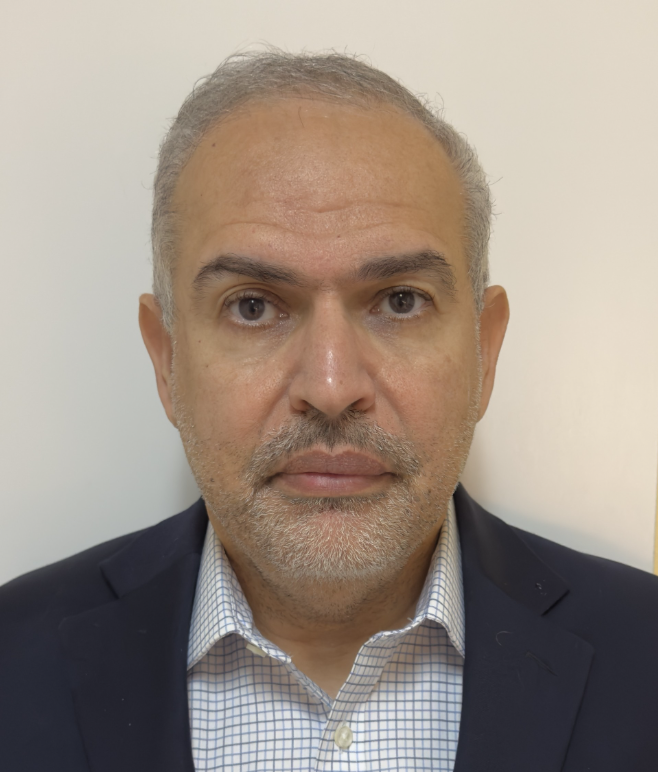}}]{\ygk{Mohamed Abdel-Aty}} \ygk{(Senior Member, IEEE) is a Pegasus Professor and a Trustee Chair Professor with the Department of Civil, Environmental and Construction Engineering, and a secondary joint appointment with the Department of Computer Science, University of Central Florida. He is a Professional Engineer. He has managed over 90 research projects. He has published more than 850 articles, of which 475 are in journals (Google Scholar citations: $>$ 40,500, H-index: 106). His main expertise and interests include ITS, simulation, CAV, and active traffic management. He is a fellow of ASCE and ITE. He is the Editor-in-Chief Emeritus of Accident Analysis and Prevention.}
\end{IEEEbiography}

\vspace{-1cm}

\begin{IEEEbiography}[{\includegraphics[width=1in,height=1.25in,clip,keepaspectratio]{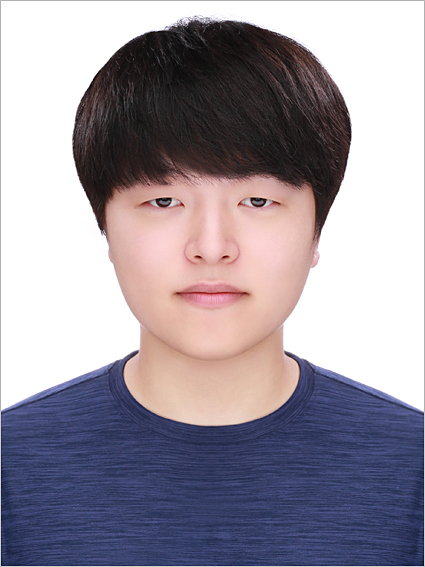}}]{Beomsik Cho} received the B.S. degree in mechanical engineering from Ajou University, Suwon, South Korea, in 2023, where he is currently pursuing the M.S. degree in AI Mobility Engineering. His research interests include artificial intelligence, domain adaptation, and autonomous vehicles.
\end{IEEEbiography}

\vspace{-1cm}

\begin{IEEEbiography}[{\includegraphics[width=1in,height=1.25in,clip,keepaspectratio]{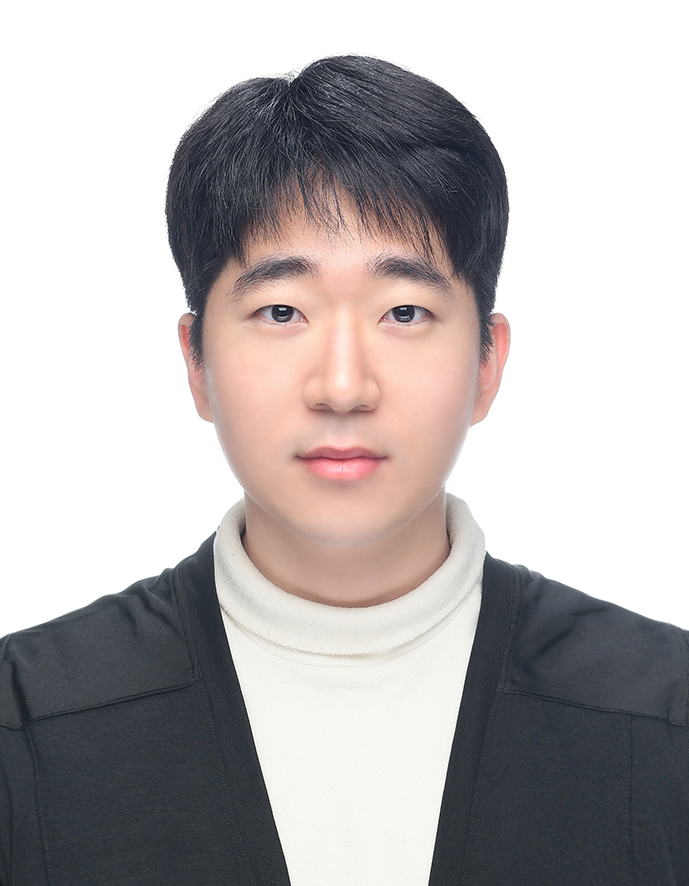}}]{Seonghoon Ryoo} received the B.S. degree in mechanical engineering from Ajou University, Suwon, South Korea, in 2024, where he is currently pursuing the M.S. degree in AI Mobility Engineering. His research interests include artificial intelligence, object detection, and sensor fusion for autonomous vehicles.
\end{IEEEbiography}

\vspace{-1cm}

\begin{IEEEbiography}[{\includegraphics[width=1in,height=1.25in,clip,keepaspectratio]{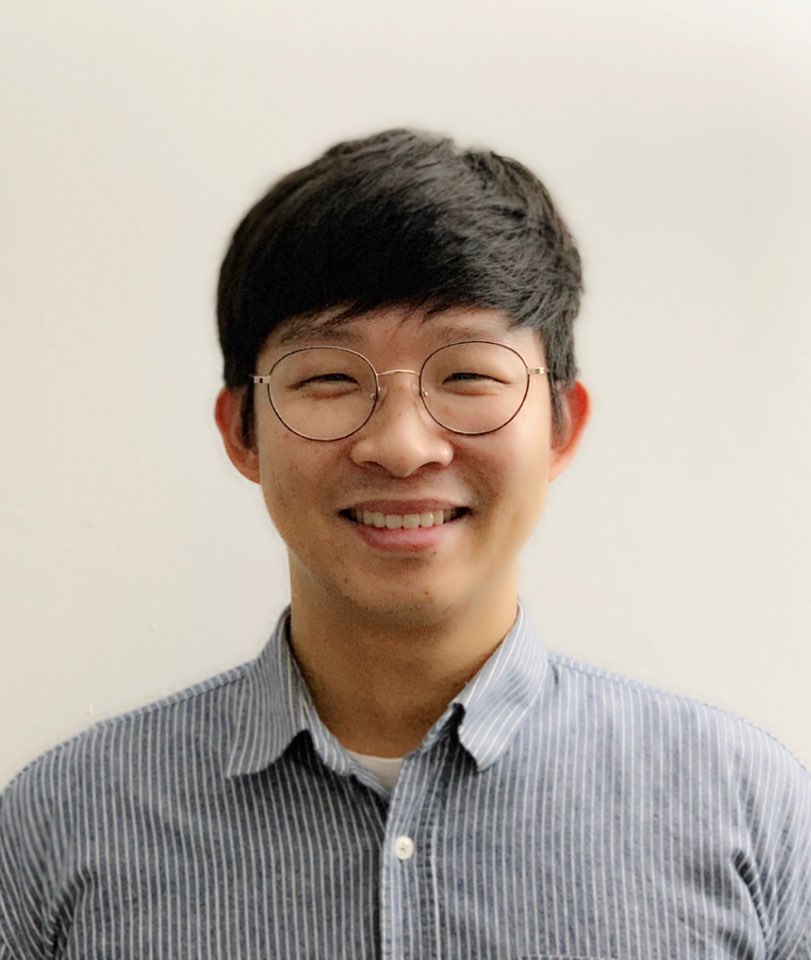}}]{Soomok Lee}  received the B.S. degree in Electrical Engineering from SungKyunKwan University, Seoul, Korea, in 2013 and the Ph.D. in Electrical and Computer Engineering from Seoul National
University, Seoul, Korea, in 2019.  Since 2022, he has been an Assistant Professor, with in the Department of  Mobility Engineering and Department of Artificial Intelligence at AJOU University, Suwon, Korea. He is currently leading the Machine Learning \& Mobility Laboratory as the Principal Investigator (PI).
From 2013 to 2017, he was a researcher at SNU Intelligent Vehicle IT Center, Seoul, Korea. From 2017 to 2022, he was a Director of SW Engineering with ThorDrive, Inc, Ohio, USA. He performed 3D multi-modal deep learning-based fusion research as a Postdoctoral researcher at Harvard Medical School / Mass General Hospital at 2022.   His current research areas include multi-modal sensor fusion, 3D computer vision, and domain adaptation.
\end{IEEEbiography}

\end{document}